\documentclass[preprint,12pt]{elsarticle}

\usepackage{amssymb}
\usepackage{amsmath}
\usepackage{booktabs}
\usepackage{tabularx}
\usepackage{array}
\usepackage{makecell}
\usepackage[table]{xcolor}
\usepackage{pifont}
\usepackage{adjustbox}
\usepackage{float}
\usepackage{xurl}
\usepackage{hyperref}

\journal{Computers and Electronics in Agriculture}

\begin{document}

\begin{frontmatter}



\title{Towards Practical Precision Agriculture: Real-Time Fruit Detection and Video Analytics on Embedded Edge Hardware} 


\author[1]{Ivica Dimitrovski} 

\author[1]{Vlatko Spasev} 

\author[1]{Ivan Kitanovski} 

\author[1]{Petre Lameski} 

\author[2]{Dane Boshev} 

\affiliation[1]{organization={Faculty of Computer Science and Engineering, University Ss Cyril and Methodius},
            addressline={ul. Rudzer Boshkovikj 16, P.O. 393}, 
            city={Skopje},
            postcode={100}, 
            country={North Macedonia}}

\affiliation[2]{organization={Faculty of Agricultural Sciences and Food, University Ss Cyril and Methodius},
            addressline={16-ta Makedonska brigada br. 3}, 
            city={Skopje},
            postcode={100}, 
            country={North Macedonia}}

\begin{abstract}
Static-image benchmarks alone do not capture the computational and temporal requirements of practical orchard video analytics. This study presents an end-to-end framework for real-time fruit detection, tracking, and counting on the NVIDIA Jetson Orin Nano Super. A lightweight YOLO26s detector is trained independently on four public datasets representing apples, mangoes, blueberries, and strawberries under a common training and evaluation protocol. This multi-dataset stage is used to assess how a common detector and deployment strategy behaves across substantially different fruit appearances, object densities, image resolutions, and orchard conditions. The resulting models are first characterized on a reference GPU and subsequently deployed on the embedded platform using PyTorch and TensorRT under FP32, FP16, and INT8 precision. The study then progresses from broad multi-crop detection benchmarking to temporal video analytics. APPLE MOTS is used for this stage because, unlike the other evaluated datasets, it additionally provides temporally ordered orchard sequences with persistent fruit identities, enabling quantitative evaluation of multi-object tracking and unique-fruit counting. The selected FP16 TensorRT detector is integrated into an NVIDIA DeepStream pipeline combining hardware-accelerated video decoding, ByteTrack multi-object tracking, and motion-aware line-crossing analytics. Across the four detection tasks, mean test mAP@50:95 ranges from 0.4957 to 0.8656. On the Jetson, TensorRT FP16 achieves 66.76--74.56 images/s at 13.41--14.98 ms prediction latency, while reducing mAP@50:95 by only 0.0020--0.0054 and gross energy consumption by approximately 64--66\% relative to PyTorch FP32. The complete detector--tracker--analytics pipeline reaches 44.96--54.11 FPS maximum throughput and sustains the configured 30-FPS input rate without output-frame loss. On the held-out orchard video sequences, Higher Order Tracking Accuracy (HOTA) ranges from 0.345 to 0.538, event-level counting F1 from 0.611 to 0.803, and relative count error from 6.2\% to 51.6\%. Performance varies substantially across acquisition geometries: near-lateral row viewing is associated with the most stable tracking and counting, whereas forward traversal between orchard rows remains association- and recall-limited despite spatially adaptive counting geometry. These results show that practical edge-based fruit monitoring requires both computationally efficient detection across diverse crop conditions and acquisition geometries that support reliable temporal association.
\end{abstract}


\end{frontmatter}

\section{Introduction}
\label{sec:introduction}

Precision agriculture increasingly relies on sensing, automation, and data-driven decision support to improve the efficiency and sustainability of crop production \cite{gebbers2010precision}. Within this context, computer vision provides a non-destructive means of obtaining information about fruit presence, spatial distribution, visible crop load, and harvesting conditions \cite{koirala2019deep}, \cite{bargoti2017deep}. Reliable fruit localization can support orchard monitoring, crop-load assessment, harvest planning, and robotic harvesting. However, agricultural vision remains substantially more difficult than object detection in controlled scenes because fruits may be small, densely clustered, partially occluded by foliage or branches, visually similar to the surrounding canopy, and observed under considerable variation in illumination, scale, and viewpoint \cite{tang2023optimization}.

Deep object detectors, and particularly models from the You Only Look Once (YOLO) family, have substantially improved the accuracy--efficiency trade-off available for fruit detection. Successful applications have been demonstrated for apples \cite{bresilla2019single}, mangoes \cite{koirala2019_v2deep,gu2024simultaneous}, citrus fruits \cite{mirhaji2021fruit}, and other crops under orchard and field conditions. More recently, YOLO26 introduced an end-to-end detector family intended to simplify the inference path and facilitate efficient deployment across heterogeneous computing platforms \cite{jocher2026ultralytics}. Nevertheless, reported fruit-detection performance remains strongly dependent on crop type, fruit scale, object density, image resolution, occlusion, and acquisition conditions \cite{vasconez2020comparison,tang2023optimization}. Evaluating a detector on a single crop or dataset therefore provides limited information about whether the same architecture and deployment strategy retain their accuracy--efficiency characteristics under substantially different agricultural imaging conditions.

A second limitation concerns the distinction between model inference and practical system execution. Many studies report image-level detection accuracy or detector inference speed, although these measurements represent only part of the computational workload encountered in an operational system. Embedded agricultural vision additionally involves image or video decoding, memory movement, preprocessing, postprocessing, temporal association, application analytics, and output handling. Previous studies have demonstrated the feasibility of embedded fruit detection and counting using resource-constrained hardware \cite{mazzia2020real,lyu2022green,gu2024simultaneous}, but the performance of the detector alone does not establish whether the complete processing chain can operate continuously under field-relevant timing and resource constraints. This distinction is particularly important for mobile and robotic platforms, where memory availability, power consumption, energy efficiency, and thermal behavior are relevant in addition to raw inference latency. Moreover, a single average throughput value can conceal timing variation or fail to distinguish maximum computational capacity from the ability to sustain a prescribed input-video rate.

Continuous video introduces a further challenge because repeated frame-level detections cannot be accumulated directly to estimate the number of unique fruits. A fruit may remain visible for many consecutive frames and must therefore be associated temporally before it can contribute reliably to a video-level count. Tracking-by-detection has consequently become an important strategy for agricultural video counting, where frame-level detections are linked into trajectories and counting is triggered by a spatial or temporal event \cite{farjon2023deep,hu2023fruit,zhang2022deep}. Alternative strategies, including trunk-assisted tracking, have also been investigated to improve counting reliability in orchard environments \cite{gao2022novel}. In all such systems, errors introduced during temporal association propagate directly to the final application: missed or prematurely terminated trajectories can produce under-counting, whereas identity switches, duplicate trajectories, or repeated events can generate incorrect counts.

The tracker itself introduces an additional accuracy--efficiency trade-off. Appearance-based association and more elaborate temporal models can improve identity continuity, but they also increase computational complexity, particularly in scenes containing many visually homogeneous fruits. ByteTrack \cite{zhang2022bytetrack} provides a comparatively lightweight alternative by associating both high- and low-confidence detections without requiring an additional appearance-embedding network. This makes it attractive for embedded deployment, but association quality remains dependent on the properties of the observed trajectories. In moving-camera orchard video, apparent fruit motion is influenced not only by detector localization and occlusion, but also by camera trajectory, viewing angle, perspective, and fruit depth. Consequently, a video pipeline that satisfies the required frame rate may still provide substantially different tracking and counting accuracy under different acquisition geometries. Recent agricultural multi-object tracking studies similarly indicate that maintaining persistent identities for dense, visually homogeneous objects remains a non-trivial problem \cite{hernandez2024multi}.

These observations motivate an evaluation that connects three aspects that are often considered separately: robustness of fruit detection across heterogeneous datasets, efficiency of deployment on embedded hardware, and temporal reliability of the resulting video-analytics pipeline. Accordingly, this study adopts a \emph{breadth-to-depth} experimental design. In the first stage, a single lightweight architecture, YOLO26s \cite{jocher2026ultralytics}, is trained independently on four public datasets: MangoYOLO for mango detection \cite{koirala2019_v2deep}, DeepBlueberry for blueberry detection \cite{gonzalez2019deepblueberry}, StrawDI\_Db1 for strawberry detection \cite{perez2020fast}, and APPLE MOTS for apple detection \cite{de2022apple}. Applying the same architecture and a common experimental protocol across all four tasks provides a controlled basis for examining dataset-dependent detection and deployment behavior without introducing variability from comparisons among unrelated detector families.

The study subsequently moves from this broad multi-crop image-level evaluation to a deeper temporal analysis. APPLE MOTS is used for this stage because it additionally provides temporally ordered orchard sequences and persistent fruit identities \cite{de2022apple}, which make it possible to evaluate multi-object tracking and unique-fruit counting quantitatively. Sequences are separated temporally between model development, tracker configuration, and final testing so that adjacent frames from the held-out video sequences do not influence parameter selection. The selected detector is integrated into an NVIDIA DeepStream pipeline comprising hardware-accelerated video decoding, TensorRT inference, ByteTrack temporal association, and motion-aware line-crossing analytics.

Rather than assuming a fixed counting direction for every video, the proposed analytics derive the dominant \emph{image-plane motion} from calibration tracklets and construct finite counting gates approximately perpendicular to the observed fruit trajectories. The estimated motion vectors describe the apparent displacement of fruits in the image and are therefore influenced by camera trajectory and perspective; they do not represent direct measurements of physical camera motion. When a single dominant motion field is insufficient, the scene can be partitioned into independent spatial motion regions, allowing separate motion directions and counting gates to be derived. This formulation is particularly relevant to forward traversal between orchard rows, where objects on opposite sides of the camera can exhibit substantially different image-plane motion.

The experimental evaluation follows the same staged progression. Image-level detection accuracy is first characterized across repeated independent training runs. Reference-GPU experiments then distinguish native PyTorch image-processing behavior from optimized TensorRT FP32 and FP16 execution. The same validation-selected checkpoints are subsequently evaluated on the NVIDIA Jetson Orin Nano Super using PyTorch and TensorRT under FP32, FP16, and INT8 precision, quantifying accuracy, latency, throughput, power consumption, energy efficiency, and thermal behavior. Finally, the selected FP16 TensorRT configuration is evaluated as part of the complete DeepStream video pipeline. Maximum-throughput runs quantify computational headroom, whereas timestamp-synchronized runs determine whether the full detector--tracker--analytics system can sustain the configured video rate. Tracking is evaluated using both detection- and association-sensitive measures, including Multiple Object Tracking Accuracy (MOTA), ID F1 score (IDF1), and Higher Order Tracking Accuracy (HOTA), while the downstream application is assessed through one-to-one event matching and sequence-level fruit-count error.

The main contributions of this work are as follows:

\begin{enumerate}

    \item \textbf{A controlled breadth-to-depth evaluation of fruit perception across heterogeneous agricultural conditions.} A common YOLO26s architecture and training protocol are applied independently to four public datasets covering mangoes, blueberries, strawberries, and apples. This multi-dataset stage characterizes how detection accuracy and deployment behavior change with crop and dataset characteristics, while the temporally annotated apple data enable a deeper subsequent evaluation of tracking and counting.

    \item \textbf{A staged, measurement-oriented analysis of embedded inference optimization.} Reference-GPU and Jetson experiments explicitly distinguish model inference latency from application-level prediction latency and evaluate PyTorch and TensorRT execution under FP32, FP16, and INT8 precision. Accuracy, throughput, power, energy consumption, and thermal behavior are measured within a consistent deployment protocol, enabling numerical precision and inference-backend effects to be evaluated separately from the later video-processing workload.

    \item \textbf{A real-time video-analytics pipeline with calibration-derived motion-aware fruit counting.} The selected TensorRT detector is integrated with hardware-accelerated decoding and ByteTrack in NVIDIA DeepStream. Dominant image-plane motion is estimated from calibration trajectories and used to construct finite direction-aware counting gates, with support for predefined spatial motion regions when a single global motion field is not representative of the scene.

    \item \textbf{A joint evaluation of tracking, counting, and real-time system behavior under different orchard acquisition geometries.} Held-out sequences with persistent identities are evaluated using MOTA, IDF1, HOTA and its detection/association decomposition, identity switches, trajectory fragmentation, event-level counting precision, recall and F1, and sequence-level count error. Maximum-throughput and timestamp-synchronized experiments are evaluated separately, allowing computational capacity to be distinguished from source-rate compliance and exposing the influence of camera viewpoint and image-plane motion on downstream tracking and counting reliability.

\end{enumerate}

The remainder of this paper is organized as follows. Section~2 reviews related work on fruit detection, tracking-based fruit counting, and embedded agricultural vision. Section~3 presents the study design, datasets, YOLO26s detector, model-conversion procedure, and motion-aware video-analytics pipeline. Section~4 describes the training protocol, reference and embedded platforms, TensorRT configurations, video-processing setup, and evaluation metrics. Section~5 reports the image-level detection, reference-GPU, embedded-deployment, and end-to-end video-analytics results. Section~6 discusses the practical implications, limitations, and threats to validity, and Section~7 concludes the paper and outlines directions for future work.

\section{Related Work}
\label{sec:related_work}

\subsection{Fruit Detection in Orchard Environments}
\label{sec:related_detection}

Vision-based fruit detection has been studied for several decades in support of crop-load estimation and robotic harvesting. Earlier reviews identified occlusion, fruit clustering, variable illumination, and sensor/viewpoint limitations as persistent challenges for machine-vision systems in orchards \cite{gongal2015sensors}. The subsequent adoption of deep learning shifted the field from hand-crafted color, texture, and shape descriptors toward learned visual representations, with comprehensive reviews documenting the rapid development of convolutional approaches for fruit detection and yield estimation \cite{koirala2019deep}.

Among the early deep-learning systems, DeepFruits demonstrated the feasibility of using Faster R-CNN for detecting multiple fruit types and showed the potential benefit of multimodal RGB and near-infrared imagery \cite{sa2016deepfruits}. Bargoti and Underwood subsequently evaluated deep convolutional detectors on apples, mangoes, and almonds under field conditions, highlighting the challenges associated with dense fruit distributions and large orchard images \cite{bargoti2017deep}. These studies established region-based convolutional detectors as an effective foundation for fruit perception in unstructured agricultural scenes.

Subsequent work increasingly investigated the trade-off between detection accuracy and computational efficiency. Vasconez \emph{et al.} compared Faster R-CNN and Single Shot Multibox Detector (SSD) across several fruit datasets and demonstrated that detector performance depends strongly on the crop, sensing conditions, and training data \cite{vasconez2020comparison}. Multi-class orchard detection using Faster R-CNN was also explored by Wan and Goudos \cite{wan2020faster}. In parallel, single-stage detectors became increasingly attractive for applications requiring real-time processing. Bresilla \emph{et al.} demonstrated real-time apple and pear detection using single-shot convolutional networks \cite{bresilla2019single}, while Koirala \emph{et al.} introduced MangoYOLO and systematically investigated the accuracy--speed trade-off for orchard fruit-load estimation \cite{koirala2019_v2deep}. Lightweight architectures such as LedNet further demonstrated that substantial model compression can be achieved while retaining useful apple-detection performance \cite{kang2020fast}.

Robust perception under dense occlusion has motivated the use of richer instance representations and multi-scale feature processing. Instance segmentation has been applied to apples and greenhouse tomatoes to separate overlapping fruit and obtain more detailed object representations \cite{kang2020fruit,afonso2020tomato}. More recently, transformer-based detectors have been introduced for agricultural detection tasks. For example, improved RT-DETR architectures have been investigated for tomato detection and phenotypic measurement under complex backgrounds \cite{gu2024tomato}. At the same time, recent lightweight YOLO-based architectures continue to target deployment efficiency \cite{alif2024yolov1}, \cite{sapkota2026comprehensive}; Faster-YOLO-AP, for example, explicitly reduces model parameters and computational cost for apple detection on edge-oriented hardware \cite{liu2024faster}.

Taken together, previous work demonstrates that high fruit-detection accuracy can be achieved using both convolutional and transformer-based architectures, but performance remains strongly dependent on crop characteristics, fruit scale, object density, occlusion, and acquisition conditions. A large proportion of recent studies therefore improve or customize a detector for a particular fruit or orchard environment. In contrast, the first stage of the present study does not propose another crop-specific detector modification. Instead, it applies the same lightweight detector architecture and common experimental protocol independently across four substantially different fruit datasets. This controlled design allows dataset-dependent accuracy and deployment behavior to be examined without simultaneously changing the detector family.

\subsection{Fruit Counting and Multi-Object Tracking}
\label{sec:related_counting}

Automatic fruit counting is closely related to pre-harvest yield estimation, for which machine vision has become one of the principal direct measurement strategies \cite{he2022fruit}. Counting from an individual image is typically performed by detecting or segmenting the visible fruit instances and using the number of predictions as the image-level count. Such approaches are straightforward but remain sensitive to missed detections, duplicate predictions, fruit occlusion, and incomplete canopy visibility. Comparative evaluations across multiple fruit types have shown that counting performance can vary substantially with the detector and acquisition conditions \cite{vasconez2020comparison}.

Video provides multiple observations of the same canopy and can expose fruit that is not visible from a single viewpoint, but it introduces the opposite problem: the same fruit may be detected repeatedly in consecutive frames. Consequently, video-based counting is commonly formulated as a tracking-by-detection problem in which frame-level detections are associated into persistent trajectories and the resulting identities are used to avoid repeated counting \cite{farjon2023deep}. The accuracy of the final count therefore depends not only on detection quality but also on identity continuity, trajectory fragmentation, re-entry after occlusion, and the strategy used to transform trajectories into counting events.

Several studies have developed fruit-specific association strategies to address these problems. Cascade-SORT combines motion and appearance cues through cascade matching and demonstrated improved robustness over SORT for camellia and apple counting \cite{he2022cascade}. Gao \emph{et al.} instead used tracked orchard trunks as a spatial reference for apple association and counting \cite{gao2022novel}. Villacrés \emph{et al.} performed a broader comparison of tracking-by-detection algorithms for apple production estimation, demonstrating that the choice of tracker and detector can substantially influence the resulting count \cite{villacres2023apple}. Related work on apple orchards has also shown that detections originating from neighboring rows or non-target regions can degrade both association and counting, motivating explicit suppression of abnormal detections \cite{wu2023ndmfcs}.

The availability of agricultural multi-object tracking datasets has enabled more direct evaluation of temporal association quality. APPLE MOTS provides temporally consistent instance identities for densely distributed apples and was introduced specifically to study detection, segmentation, and tracking of visually homogeneous objects \cite{de2022apple}. Hernandez and Medeiros later used an adapted version of this benchmark to investigate spatial association with Vision Transformers, showing that explicit spatial information can improve agricultural multi-object tracking under moving-camera conditions \cite{hernandez2024multi}. For citrus orchards, Santos \emph{et al.} introduced the MOrangeT dataset and incorporated three-dimensional fruit relocalization to handle occlusion, disappearance, and re-entry during video-based counting \cite{santos2024multiple}. These studies emphasize that the agricultural tracking problem differs from conventional pedestrian tracking because fruits are visually similar and often physically stationary, while most of the observed image motion is induced by movement of the camera.

Recent work has increasingly integrated tracking and application-specific counting logic. StraTracker combines detection, motion and appearance association, adaptive Kalman filtering, and dual-region counting for strawberries \cite{an2024stratracker}. A recent row-scale kiwifruit system similarly combines fruit tracking with a verification mechanism and dynamically adapts the detection region using orchard support posts \cite{zhang2025row}. The latter is particularly indicative of a broader trend: orchard structure and acquisition geometry can be exploited explicitly rather than treating every frame as an unconstrained generic tracking scene.

These studies establish tracking-by-detection as a practical basis for video-based fruit counting, but they also show that strong detector accuracy does not guarantee reliable temporal counting. Association errors, trajectory fragmentation, camera-induced motion, and the definition of the counting region can dominate downstream performance. The present study therefore evaluates tracking and counting separately: persistent ground-truth identities are used to quantify temporal association, while predicted and reference line-crossing events are evaluated independently. Moreover, rather than assuming a single manually fixed counting direction, the deployed analytics derive dominant image-plane motion from calibration trajectories and use it to construct finite motion-aware counting gates, including separate spatial motion regions when a single motion field is insufficient.

\subsection{Edge AI and Embedded Orchard Vision}
\label{sec:related_edge}
Field-deployable agricultural vision systems must satisfy computational constraints that are generally absent from offline detector benchmarks. On-device inference can reduce reliance on network connectivity and support mobile sensing and robotic operation, but embedded platforms impose limits on compute capacity, memory, energy consumption, and thermal dissipation. Consequently, considerable research has focused on lightweight detector architectures and model compression for agricultural edge computing.

Mazzia \emph{et al.} developed an embedded apple-detection system based on a customized YOLOv3-tiny architecture and evaluated hardware-accelerated execution on resource-constrained platforms, including the NVIDIA Jetson Nano \cite{mazzia2020real}. LedNet similarly targeted real-time apple detection through a compact network architecture \cite{kang2020fast}. For mango applications, MangoYOLO demonstrated an early accuracy--speed trade-off oriented toward practical orchard monitoring \cite{koirala2019_v2deep}. Lyu \emph{et al.} subsequently integrated a lightweight YOLOv5-based citrus detector and counting method into an AI edge system \cite{lyu2022green}.

More recent studies have moved to newer embedded GPU platforms and increasingly combine lightweight networks with deployment-level optimization. Gu \emph{et al.} deployed an improved YOLOv8 model for simultaneous mango and fruiting-stem detection on the NVIDIA Jetson Orin Nano \cite{gu2024simultaneous}. Faster-YOLO-AP similarly targets computationally efficient apple detection through architectural simplification \cite{liu2024faster}. Li \emph{et al.} recently combined a lightweight passion-fruit detector with TensorRT optimization on the Jetson Orin Nano, demonstrating the continuing shift from workstation-only evaluation toward field-oriented edge execution \cite{li2025optimizing}.

A parallel trend is the development of more integrated orchard sensing systems. Agrosense v2, for example, combines temporal tree identification, canopy characterization, fruit detection, and geo-referenced aggregation in a real-time orchard-monitoring framework \cite{liu2026agrosense}. Such system-level approaches are important because practical agricultural perception requires multiple processing stages to operate continuously rather than treating detector inference as an isolated task.

Nevertheless, many embedded fruit-perception studies still characterize real-time feasibility primarily through detector-level inference speed or frame rate. These measurements do not necessarily represent the computational behavior of a complete video system, where hardware decoding, preprocessing, inference, temporal association, analytics, memory movement, and output synchronization execute together. They also provide limited information about whether reduced numerical precision changes detection accuracy or whether higher throughput translates into lower energy consumption under sustained operation.

The present work addresses this distinction through a staged deployment evaluation. Reference-GPU measurements first separate model inference, application-level prediction, and image I/O effects. The same validation-selected detector checkpoints are then evaluated on the NVIDIA Jetson Orin Nano Super using PyTorch and TensorRT under FP32, FP16, and INT8 precision, with accuracy, latency, throughput, power, energy, and temperature measured explicitly. Finally, the selected TensorRT configuration is incorporated into the complete hardware-decoded detector--tracker--analytics pipeline. Maximum-throughput and timestamp-synchronized experiments are treated separately so that computational capacity can be distinguished from the ability to sustain the configured video rate. This system-level evaluation complements prior detector-centric edge studies by relating embedded efficiency directly to tracking and fruit-counting performance.

\section{Materials and Methods}
\label{sec:materials_methods}

\subsection{Study Design and System Overview}
\label{sec:study_design}

The study follows a staged \emph{breadth-to-depth} design that progressively moves from controlled multi-crop fruit detection to embedded inference optimization and, finally, complete temporal video analytics. The objective is not to compare a large number of detector architectures, but to keep the perception model fixed and examine how detection accuracy, computational behavior, and temporal reliability change with dataset characteristics, deployment backend, numerical precision, and orchard acquisition geometry.

\begin{figure}[t]
\centering
\includegraphics[width=1.0\linewidth]{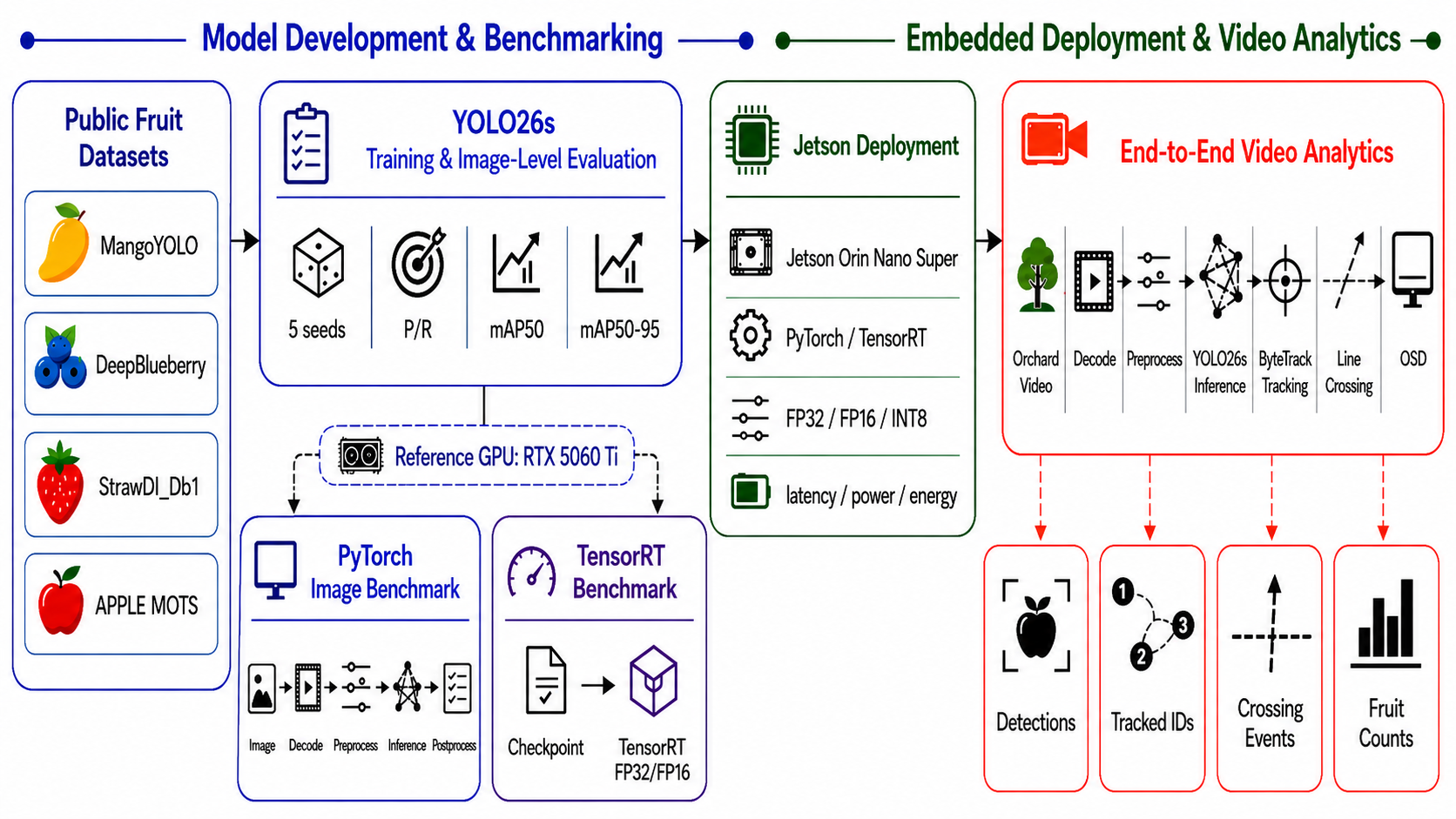}
\caption{Overview of the breadth-to-depth experimental workflow. A common YOLO26s detector is first evaluated across four fruit datasets, followed by reference-GPU characterization on an NVIDIA GeForce RTX 5060 Ti with 16~GB of graphics memory and embedded PyTorch/TensorRT optimization on an NVIDIA Jetson Orin Nano Super with 8~GB of unified memory. The selected FP16 TensorRT configuration is then integrated into a DeepStream-based video pipeline with ByteTrack association and calibration-derived motion-aware fruit counting.}
\label{fig:system_architecture}
\end{figure}

As illustrated in Fig.~\ref{fig:system_architecture}, the first stage evaluates a single lightweight detector architecture, YOLO26s, independently on four public fruit datasets representing mangoes, blueberries, strawberries, and apples. The datasets differ substantially in fruit size, object density, occlusion, native image resolution, background complexity, and acquisition viewpoint. Using the same detector architecture and common experimental protocol across all four tasks limits architectural variability and provides a controlled basis for studying dataset-dependent detection and deployment behavior.

The second stage characterizes computational execution independently from training variability. Detection is first evaluated on a reference desktop GPU, where native PyTorch image-processing behavior is separated from optimized TensorRT FP32 and FP16 execution. The same validation-selected checkpoints are then transferred to the NVIDIA Jetson Orin Nano Super and evaluated using PyTorch and TensorRT under FP32, FP16, and INT8 precision. This stage quantifies the effects of inference backend and numerical representation on accuracy, latency, throughput, power, energy efficiency, and thermal behavior.

The final stage moves from broad multi-crop detection benchmarking to a deeper temporal evaluation. APPLE MOTS is used for this stage because, in addition to frame-level fruit annotations, it provides temporally ordered orchard sequences and persistent instance identities. These annotations make it possible to evaluate multi-object tracking and unique-fruit counting quantitatively. The selected FP16 TensorRT detector is integrated into an NVIDIA DeepStream pipeline comprising hardware-accelerated video decoding, TensorRT inference, ByteTrack multi-object tracking, and motion-aware line-crossing analytics.

Two complementary runtime modes are used for the complete video system. Maximum-throughput processing disables source-rate synchronization and measures the computational capacity of the detector--tracker--analytics chain. Timestamp-synchronized processing preserves the configured video cadence and evaluates whether the complete pipeline can operate continuously at the 30-FPS source rate. Tracking and counting accuracy are evaluated separately from runtime so that computational feasibility can be distinguished from temporal perception quality.

\subsection{Datasets and Data Preparation}
\label{sec:datasets}

Four public fruit datasets are used to represent substantially different agricultural detection conditions: MangoYOLO for mango detection \cite{koirala2019_v2deep}, DeepBlueberry for blueberry detection \cite{gonzalez2019deepblueberry}, StrawDI\_Db1 for strawberry detection \cite{perez2020fast}, and APPLE MOTS for apple detection and temporal video analysis \cite{de2022apple}. The first three datasets contribute to the multi-crop image-level benchmark, whereas APPLE MOTS additionally supports multi-object tracking and fruit-counting evaluation. Table~\ref{tab:datasets} summarizes their principal characteristics.

For MangoYOLO and StrawDI\_Db1, the predefined training, validation, and test partitions distributed with the datasets are retained. DeepBlueberry does not provide an official partition; its images are therefore divided once into fixed training, validation, and test subsets using a 70\%/15\%/15\% split. The resulting partition remains unchanged for every experimental run so that run-to-run variation originates from model training rather than from changes in the underlying samples.

APPLE MOTS requires a different protocol because adjacent frames are strongly temporally correlated. The data are therefore separated at the sequence level rather than by randomly partitioning individual frames. Sequences 0000--0004 are used for detector training, sequence 0005 is reserved for model validation and video-pipeline parameter selection, and sequences 0006--0008 are retained exclusively for final testing. Consequently, temporally adjacent frames from a held-out sequence cannot enter either detector training or tracker configuration.

APPLE MOTS provides instance segmentation masks together with persistent object identifiers. For detector training and image-level evaluation, each instance mask is converted to its axis-aligned bounding box while retaining the corresponding image geometry. For temporal evaluation, the original persistent identities are preserved and used as tracking ground truth. Bottom-center trajectories derived from the annotated boxes are additionally used to generate reference crossing events for the counting experiment.

Sequences 0010--0012 are not included in the primary benchmark because the released RGB images contain visually overlaid ignore regions. Incorporating these sequences into the same YOLO detection protocol would therefore require additional masking or sequence-specific preprocessing not applied to the remaining data. Restricting the benchmark to sequences 0000--0008 preserves a consistent image representation across detector training, static-image evaluation, and temporal analysis.

For image-level testing, the frames of sequences 0006--0008 are treated as independent images, consistent with the other detection datasets. For the video experiment, exactly the same held-out frames are restored to their original temporal order and encoded as 30-FPS video streams. Thus, the apple detection and temporal experiments use the same unseen image content, while the latter additionally exploits the temporal identities required for tracking and counting.

\begin{table}[H]
\centering
\caption{Summary and experimental role of the fruit datasets.}
\label{tab:datasets}

\small
\setlength{\tabcolsep}{3.5pt}
\renewcommand{\arraystretch}{1.10}

\begin{adjustbox}{max width=\linewidth}
\begin{tabular}{lcccc}
\toprule
\textbf{Dataset} &
\textbf{Crop} &
\textbf{\#Images} &
\textbf{\#Instances} &
\textbf{Experimental role} \\
\midrule

MangoYOLO \cite{koirala2019_v2deep}
& Mango
& 1,730
& 15,281
& Detection \\

DeepBlueberry \cite{gonzalez2019deepblueberry}
& Blueberry
& 305
& 10,132
& Detection \\

StrawDI\_Db1 \cite{perez2020fast}
& Strawberry
& 3,100
& 17,938
& Detection \\

APPLE MOTS \cite{de2022apple}
& Apple
& 1,673
& 85,958
& Detection + video analytics \\

\bottomrule
\end{tabular}
\end{adjustbox}
\end{table}

Because native image resolution is relevant to the runtime experiments, the source dimensions are retained during image loading. MangoYOLO images are $612\times512$ pixels, StrawDI\_Db1 images are $1008\times756$ pixels, APPLE MOTS frames are $1296\times972$ pixels, and DeepBlueberry contains images at $1920\times1080$ and $2448\times3264$ pixels. All images are subsequently transformed to the fixed network input resolution described in Section~\ref{sec:experimental_setup}.

\begin{figure}[H]
\centering
\includegraphics[
    width=\linewidth,
    keepaspectratio
]{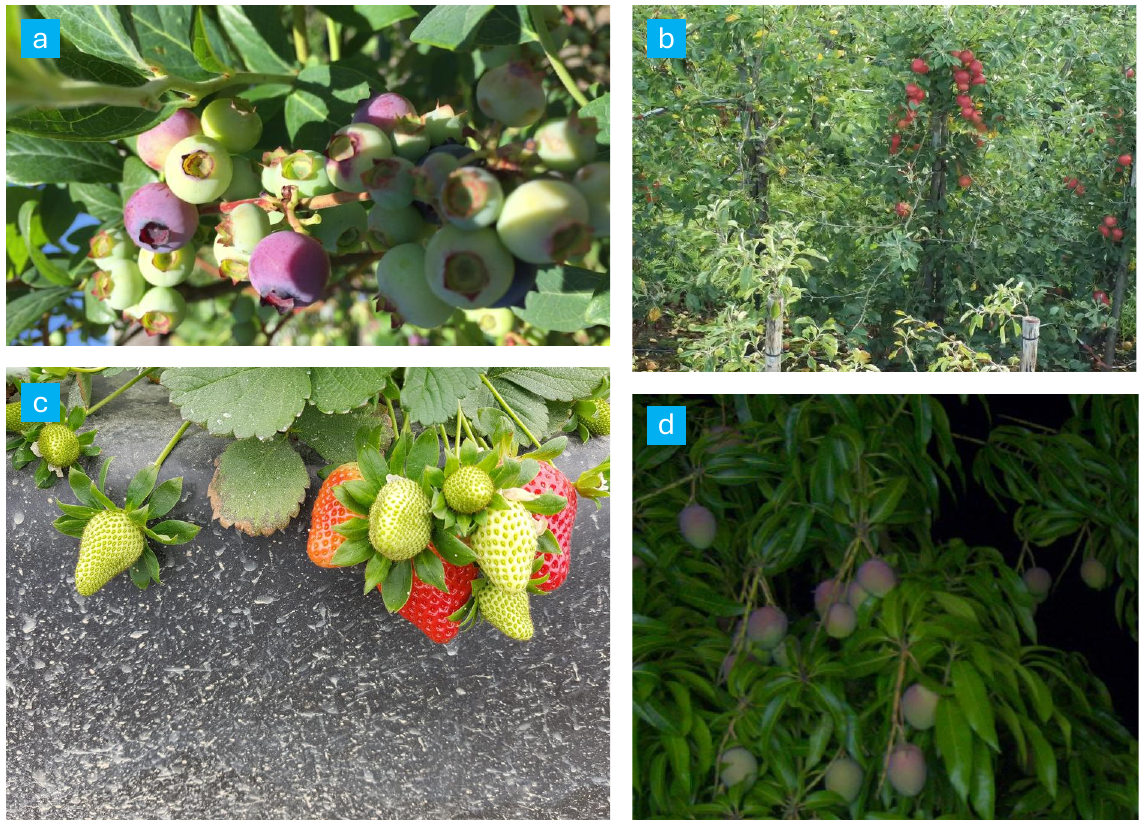}
\caption{Representative examples from the four fruit datasets: (a) DeepBlueberry, (b) APPLE MOTS, (c) StrawDI\_Db1, and (d) MangoYOLO. The datasets exhibit substantial variation in fruit scale, object density, occlusion, native image resolution, background complexity, and acquisition viewpoint.}
\label{fig:dataset_examples}
\end{figure}

Representative samples are shown in Fig.~\ref{fig:dataset_examples}. The examples illustrate why the four datasets are treated as complementary detection environments rather than as a single pooled training dataset.

\subsection{YOLO26s Fruit Detector}
\label{sec:yolo26}

Fruit detection is performed using YOLO26s, the small variant of the Ultralytics YOLO26 family \cite{jocher2026ultralytics}. A single detector capacity is intentionally retained across all datasets so that differences in accuracy and computational behavior are not confounded by simultaneously changing model size or detector family.

YOLO26s follows the multi-scale organization characteristic of modern YOLO detectors. A convolutional backbone extracts hierarchical visual features, which are aggregated through a Feature Pyramid Network--Path Aggregation
Network (FPN--PAN) neck before prediction at multiple spatial resolutions. The backbone and feature-fusion stages employ C3k2, Spatial Pyramid Pooling--Fast (SPPF), and Cross-Stage Partial Spatial Attention (C2PSA) components. Detection is performed using an anchor-free decoupled head.

A deployment-relevant characteristic of YOLO26 is its end-to-end detection path. In contrast to earlier Ultralytics architectures that employ Distribution Focal Loss (DFL) for bounding-box regression and require a separate Non-Maximum Suppression (NMS) stage at inference, YOLO26 directly regresses box coordinates and employs dual one-to-many and one-to-one prediction branches. The one-to-many branch provides dense supervision during training, whereas the one-to-one branch is optimized to return non-duplicated detections and is retained by the end-to-end inference/export path. This NMS-free output formulation is particularly useful for TensorRT and DeepStream deployment because the final detector graph requires less external postprocessing.

A separate YOLO26s model is fine-tuned for each dataset. No cross-dataset training or transfer evaluation is performed; each dataset constitutes an independent within-dataset detection task. Detector architecture, input resolution, training policy, and evaluation procedure are otherwise held constant. Repeated training is used to quantify training variability, whereas all subsequent runtime comparisons use one fixed checkpoint per dataset selected only according to validation performance. The complete training and checkpoint-selection procedure is given in Section~\ref{sec:experimental_setup}.

\subsection{Model Conversion and Embedded Deployment}
\label{sec:model_deployment}

Following detector training, the validation-selected PyTorch checkpoints are converted for optimized NVIDIA execution. Export proceeds through the Open Neural Network Exchange (ONNX) representation, after which TensorRT engines are constructed directly on the hardware on which they are evaluated. Engines are therefore not transferred between the reference GPU and the Jetson platform, avoiding hardware-specific TensorRT compilation differences.

Static batch-one engines are used with the same network input dimensions and end-to-end YOLO26 output representation as the corresponding PyTorch models. Three numerical representations are evaluated on the Jetson: 32-bit floating point (FP32), 16-bit floating point (FP16), and 8-bit integer (INT8). FP32 provides the numerical reference for TensorRT execution, FP16 exploits the native reduced-precision capabilities of the GPU, and INT8 uses post-training quantization with representative calibration samples drawn exclusively from the corresponding training partition.

Detection accuracy is re-evaluated after engine construction rather than assuming numerical equivalence with the source PyTorch checkpoint. This makes it possible to quantify the accuracy cost of optimization separately from its computational benefit. In addition to model-execution latency, the deployment analysis considers synchronized application-level prediction latency, throughput, module power, energy per processed image, and temperature.

The TensorRT FP16 configuration is subsequently used as the detector stage of the video experiment. The engine is loaded by the DeepStream \texttt{nvinfer} component so that each video frame is decoded and inferred once before detector metadata are passed directly to the temporal association stage. Exact engine construction, INT8 calibration, runtime, and telemetry settings are reported in Section~\ref{sec:experimental_setup}.

\subsection{Video Analytics, Tracking, and Motion-Aware Counting}
\label{sec:video_analytics}

\subsubsection{DeepStream Processing and ByteTrack Association}
\label{sec:video_pipeline}

The proposed video-analytics system operates on temporally ordered orchard video using an NVIDIA DeepStream pipeline. Input video is hardware decoded, passed through \texttt{nvstreammux}, and processed by the selected FP16 TensorRT YOLO26s detector through \texttt{nvinfer}. Detector bounding boxes, class identifiers, and confidence scores are propagated through DeepStream object metadata to the temporal association stage. The detector is therefore executed only once per frame, and the resulting metadata are reused directly for tracking and subsequent counting analytics.

Temporal association is performed using ByteTrack \cite{zhang2022bytetrack}. ByteTrack follows a tracking-by-detection strategy in which high-confidence detections are first associated with active trajectories, after which lower-confidence detections are considered in a second association stage to recover objects whose detector confidence has temporarily decreased. This mechanism is particularly relevant in orchard video, where partial occlusion, motion blur, changing viewpoint, and dense fruit arrangements can cause short-term confidence fluctuations.

The implementation does not use an additional appearance-embedding network or image-based global-motion-compensation module. Tracking therefore relies primarily on the spatial and temporal consistency of detector outputs. This choice limits the additional computational cost of temporal association and is consistent with the objective of maintaining real-time execution on resource-constrained embedded hardware.

The resulting persistent track identities provide the temporal representation used by the downstream counting stage. Rather than accumulating detections independently across frames, each tracked fruit contributes a trajectory in image coordinates. Counting events are subsequently derived from these trajectories using the calibration-based motion and gate construction described in the following subsections.

\subsubsection{Calibration-derived image-plane motion}
\label{sec:motion_calibration}

A fixed manually specified counting direction is not appropriate for all moving-camera orchard videos because apparent fruit motion depends on the camera trajectory, viewing angle, depth, and perspective. The counting geometry is therefore estimated from an initial unlabeled calibration interval of each video using only ByteTrack predictions. Ground-truth boxes, identities, and reference crossing events are not used during geometry estimation.

For each tracked identity $i$, the bottom-center position of its bounding box is represented as

\begin{equation}
    \mathbf{p}_{i,t} =
    \begin{bmatrix}
        x_{i,t}\\
        y_{i,t}
    \end{bmatrix}.
\end{equation}

Only calibration tracklets satisfying minimum duration, observation-count, and displacement criteria are retained. To reduce sensitivity to frame-level localization jitter, start and end positions are estimated from small endpoint windows of each tracklet. The resulting displacement is

\begin{equation}
    \Delta\mathbf{p}_i =
    \mathbf{p}^{\,\mathrm{end}}_i -
    \mathbf{p}^{\,\mathrm{start}}_i,
\end{equation}

and the corresponding unit motion vector is

\begin{equation}
    \mathbf{u}_i =
    \frac{\Delta\mathbf{p}_i}
         {\left\|\Delta\mathbf{p}_i\right\|_2}.
\end{equation}

Tracklets with larger spatial displacement and longer temporal support are expected to provide more reliable estimates of image-plane motion. To avoid mixing quantities expressed in pixels and frames, both terms are normalized before weighting. Let $f_i^{\mathrm{start}}$ and $f_i^{\mathrm{end}}$ denote the first and last frame indices of tracklet $i$, respectively, and define its temporal span as $\Delta f_i = f_i^{\mathrm{end}} - f_i^{\mathrm{start}}$. The normalized spatial displacement and temporal support of tracklet $i$ are defined as

\begin{equation}
    d_i =
    \frac{
        \left\|\Delta\mathbf{p}_i\right\|_2
    }{
        \sqrt{W^2+H^2}
    },
    \qquad
    t_i =
    \frac{
        \Delta f_i
    }{
        F_{\mathrm{cal}}
    },
\end{equation}

where $W$ and $H$ denote the image dimensions and $F_{\mathrm{cal}}$ is the length of the calibration interval in frames. The reliability weight is then

\begin{equation}
    w_i = \sqrt{d_it_i}.
\end{equation}

The square-root transformation limits the influence of exceptionally long or high-displacement trajectories while still assigning greater support to tracklets that provide stronger spatial and temporal evidence. Because the
normalization factors are common to all tracklets within a sequence, this formulation preserves the relative weighting used for dominant-direction estimation.

A robust dominant direction is estimated in two steps. First, each observed unit vector $\mathbf{u}_j$ is considered in turn as a candidate direction $\mathbf{c}=\mathbf{u}_j$, so that $\|\mathbf{c}\|_2=1$, and its weighted support from directionally compatible tracklets is evaluated. Given candidate $\mathbf{c}$ and cosine-compatibility threshold $\tau$, the support is

\begin{equation}
    S(\mathbf{c}) =
    \sum_{i:\,\mathbf{c}^{T}\mathbf{u}_i\geq\tau}
    w_i\,\mathbf{c}^{T}\mathbf{u}_i.
\end{equation}

The candidate receiving the strongest support defines the inlier set $\mathcal{I}$. The final dominant image-plane motion vector is then obtained from the normalized weighted vector sum

\begin{equation}
    \hat{\mathbf{v}} =
    \frac{
        \sum_{i\in\mathcal{I}}w_i\mathbf{u}_i
    }{
        \left\|
        \sum_{i\in\mathcal{I}}w_i\mathbf{u}_i
        \right\|_2
    }.
\end{equation}

Directional concentration is retained as a calibration diagnostic,

\begin{equation}
    R =
    \frac{
        \left\|
        \sum_{i\in\mathcal{I}}w_i\mathbf{u}_i
        \right\|_2
    }{
        \sum_{i\in\mathcal{I}}w_i
    },
\end{equation}

where values close to one indicate strongly aligned image-plane trajectories. By construction, $0 \leq R \leq 1$, where values close to one indicate a highly concentrated set of motion directions and lower values indicate greater directional dispersion. An additional concentration value is computed before robust inlier filtering to expose scenes containing more than one substantial motion regime.

Importantly, $\hat{\mathbf{v}}$ describes the dominant motion of fruits in the \emph{image plane}; it is not an estimate of the physical camera-motion vector. Since fruits are approximately stationary in the orchard, their apparent motion is induced primarily by camera movement but is additionally affected by perspective and object depth.

\subsubsection{Motion-aware counting-gate placement}
\label{sec:gate_placement}

The orientation and location of the counting gate are both derived from the calibration tracklets. Let $\hat{\mathbf{v}}$ denote the dominant image-plane motion direction. Each reliable tracklet is projected onto this motion axis,

\begin{equation}
    s = \mathbf{p}^{T}\hat{\mathbf{v}}.
\end{equation}

Its start and end positions therefore define an interval $[s_i^{\min},s_i^{\max}]$ along the dominant motion direction. A small margin is removed from both ends of each interval, yielding the trimmed interval $[\tilde{s}_i^{\min},\tilde{s}_i^{\max}]$, so that a counting gate is not preferentially placed where trajectories are first initialized or disappear from the scene.

Candidate gate coordinates are then evaluated according to how many trimmed inlier-tracklet intervals they intersect. The selected coordinate $s^{*}$ maximizes this coverage,

\begin{equation}
s^{*}
=
\arg\max_{s}
\sum_{i \in \mathcal{I}}
\mathbf{1}
\left\{
s \in
\left[
\tilde{s}_{i}^{\min},
\tilde{s}_{i}^{\max}
\right]
\right\},
\end{equation}

with ties resolved using robust weighted support and proximity to the median tracklet midpoint. Here, $\mathbf{1}\{\cdot\}$ denotes the indicator function. The counting line is defined by

\begin{equation}
    \mathcal{L} =
    \left\{
        \mathbf{p}:
        \mathbf{p}^{T}\hat{\mathbf{v}} = s^{*}
    \right\},
\end{equation}

and therefore has tangent direction

\begin{equation}
    \mathbf{g} =
    \begin{bmatrix}
        -\hat{v}_y\\
        \hat{v}_x
    \end{bmatrix},
\end{equation}

which is perpendicular to the estimated fruit motion. The infinite line is finally clipped to the valid image or region boundary to obtain the finite counting gate used by the analytics.

The procedure supports multiple non-overlapping spatial motion regions when a single dominant motion field does not adequately represent the scene. In such cases, the image is partitioned into predefined spatial regions, and dominant image-plane motion and gate placement are estimated independently within each region using only the calibration tracklets assigned to that region. This allows the counting geometry to adapt to scenes in which objects exhibit different apparent motion patterns across the field of view, such as when multiple crop rows or strongly perspective-dependent regions are visible simultaneously.

The spatial partition itself is defined independently of ground-truth identities and reference crossing annotations. Once the calibration step is completed, the derived motion directions and finite counting gates are frozen and used unchanged during online counting. For evaluation, the same frozen spatial geometry and direction definitions are applied to the reference trajectories so that predicted and ground-truth events are compared under an identical spatial event definition.

\subsubsection{Trajectory crossing and unique-fruit counting}
\label{sec:crossing_logic}

Fruit counting is performed from ByteTrack identities rather than from the number of frame-level detector outputs. For each active identity, the bottom-center trajectory is maintained over a bounded recent history. Intersection with a finite counting gate is evaluated using consecutive trajectory observations. Segments separated by an excessive temporal gap are not interpolated, preventing a tracker reappearance after a long interruption from being interpreted automatically as a valid physical crossing.

A geometric intersection alone is insufficient to produce an event. The recent displacement of the trajectory must also be directionally compatible with the calibrated motion vector of the corresponding gate. Direction is estimated over multiple recent observations rather than from only the single segment intersecting the line, reducing sensitivity to localization jitter and short oscillations near the gate. If the longer displacement is too small to define a reliable direction, progressively more local trajectory segments are considered.

For trajectory displacement $\Delta\mathbf{q}$ and calibrated gate direction $\hat{\mathbf{v}}$, directional agreement is measured by

\begin{equation}
    \rho =
    \frac{
        \Delta\mathbf{q}^{T}\hat{\mathbf{v}}
    }{
        \left\|\Delta\mathbf{q}\right\|_2
    }.
\end{equation}

Since $\|\hat{\mathbf{v}}\|_2=1$, $\rho$ is the cosine of the angle between the recent trajectory displacement $\Delta\mathbf{q}$ and the calibrated motion direction $\hat{\mathbf{v}}$, and therefore satisfies $-1 \leq \rho \leq 1$. Only crossings with motion aligned with the permitted direction are retained. Each ByteTrack identity can contribute at most one physical counting event across all gates. In a multi-region configuration this global uniqueness constraint prevents a trajectory near a region boundary from being counted independently by multiple gates.

For quantitative evaluation, the same crossing formulation can be applied to reference trajectories to generate ground-truth events under the frozen counting geometry. Predicted and reference event generation therefore share the same spatial gate definition and permitted motion direction, ensuring that both are evaluated against an identical event geometry. Accordingly, the reference count represents the number of ground-truth fruit trajectories that generate a valid crossing event under this frozen counting geometry; it should not be interpreted as the total number of unique fruits visible in the sequence.

Trajectory continuity may nevertheless be handled using different maximum-gap criteria for predicted and reference trajectories, since tracker outputs and annotated trajectories can exhibit different continuity characteristics. The dataset-specific gap settings, one-to-one event-matching procedure, and tracking/counting evaluation metrics are specified in Section~\ref{sec:experimental_setup}.

\section{Experimental Setup}
\label{sec:experimental_setup}

\subsection{Detector Training and Checkpoint Selection}
\label{sec:training_setup}

YOLO26s is trained independently for each dataset from the COCO-pretrained \texttt{yolo26s.pt} checkpoint. Five independent runs are performed using seeds $\{0,1,2,3,4\}$. Each run uses 100 epochs, a batch size of 8, and a fixed network input resolution of $640\times640$ pixels. Automatic mixed-precision training is enabled, optimizer selection follows the Ultralytics automatic policy, cosine learning-rate scheduling is disabled, and mosaic augmentation is disabled during the final 10 epochs. Parameters not explicitly overridden retain the defaults of the fixed Ultralytics version used throughout the experiments.

Validation is performed during training, and the checkpoint obtaining the best validation performance is retained independently for each seed. Each retained checkpoint is evaluated once on the corresponding fixed test partition using the same $640\times640$ input resolution, a batch size of 8, and a maximum of 1,000 detections per image to accommodate densely populated fruit scenes. Precision, recall, mAP@50, and mAP@50:95 are subsequently aggregated across the five runs as mean and sample standard deviation.

The repeated-run experiment is used only to characterize detector accuracy and training variability. For all subsequent runtime and deployment experiments, one checkpoint per dataset is selected exclusively according to validation performance and then frozen. The same checkpoint is used for all PyTorch and TensorRT precision modes for the corresponding dataset, ensuring that differences observed in the runtime experiments originate from inference backend and numerical precision rather than from independently trained model weights.

\subsection{Reference-GPU Runtime Evaluation}
\label{sec:reference_gpu_setup}

Training and reference-platform runtime measurements are performed on an NVIDIA GeForce RTX 5060 Ti with 16~GB of graphics memory. Two complementary protocols are used to distinguish complete native-image processing through PyTorch from optimized TensorRT execution.

For the PyTorch protocol, the complete test partition is processed sequentially at batch size one. Source images are read and decoded at their native resolution before being transformed to the fixed $640\times640$ network input. Inference uses a confidence threshold of 0.25 and a maximum of 1,000 detections per image. The first 30 predictions are treated as warm-up and excluded from runtime statistics. CUDA execution is explicitly synchronized around measured operations to prevent asynchronous kernel execution from biasing wall-clock measurements.

Three timing scopes are retained. \emph{Inference latency} denotes the backend-reported execution time of the detector itself. \emph{Prediction latency} is the CUDA-synchronized wall-clock duration from preprocessing of an image already available in memory through model execution, post-processing, and construction of the final detections. \emph{End-to-end (E2E) latency} additionally includes native-image file reading and decoding and is therefore reported for the sequential PyTorch image-processing benchmark.

For the TensorRT protocol, the validation-selected checkpoint for each dataset is independently exported to static FP32 and FP16 engines on the RTX 5060 Ti. Engines use batch-one $640\times640$ input and the native YOLO26 end-to-end output path. The confidence threshold is 0.25 and at most 300 detections are returned per image. Up to 200 images from the test partition are decoded and preloaded into system memory before measurement so that TensorRT timing does not include disk I/O or source-image decoding.

Each TensorRT configuration receives 100 warm-up predictions followed by five measured rounds. CUDA synchronization is applied around every measured prediction. GPU utilization, graphics-memory consumption, board power, and temperature are sampled with \texttt{nvidia-smi} at 100-ms intervals. A configuration is started only after the GPU temperature has returned to approximately $50\,^{\circ}\mathrm{C}$, reducing temperature-dependent variation among measurements. TensorRT throughput is derived from synchronized prediction latency, while energy per processed image is obtained from mean measured GPU power and achieved throughput.

\subsection{Embedded Inference and TensorRT Quantization}
\label{sec:embedded_setup}

Embedded experiments are performed on the NVIDIA Jetson Orin Nano Super. The device operates in the MAXN SUPER power mode configured through \texttt{nvpmodel}. CPU, GPU, and memory-controller clocks are locked using \texttt{jetson\_clocks}, active fan cooling is enabled, and the same power, clock, and cooling configuration is retained for all experiments. The software environment comprises JetPack~\textit{6.2.1}, L4T~\textit{36.4.7}, CUDA~\textit{12.9}, cuDNN~\textit{9.20}, TensorRT~\textit{10.7.0}, PyTorch~\textit{2.8.0}, Ultralytics~\textit{8.4.117}, and NVIDIA DeepStream~\textit{7.1}.

Five execution configurations are evaluated for each frozen dataset-specific checkpoint: PyTorch FP32, PyTorch FP16, TensorRT FP32, TensorRT FP16, and TensorRT INT8. All configurations use batch-one inference, a $640\times640$ network input, and a confidence threshold of 0.25. TensorRT engines are constructed directly on the Jetson rather than transferred from the reference GPU. Static input dimensions, graph simplification, the native end-to-end YOLO26 output path, and a maximum of 300 detections per image are used consistently across TensorRT precision modes.

INT8 engines are generated using post-training quantization. Calibration images are drawn exclusively from the training partition of the corresponding dataset; validation and test images are never used for quantization calibration. The complete available training split is supplied to the calibration procedure. Under the TensorRT~10.7.0 GPU calibration path used in the experiments, Ultralytics constructs an \texttt{IInt8Calibrator}-based \texttt{EngineCalibrator} using MINMAX calibration. The resulting INT8 engine is evaluated on the unchanged test partition using the same detection protocol as the corresponding FP32 and FP16 configurations.

For runtime measurement, up to 200 test images are decoded and preloaded into system RAM before timing. Each configuration receives 50 synchronized warm-up predictions followed by 60~s of continuous sequential batch-one inference. CUDA synchronization is applied immediately before and after each measured prediction. Preprocessing, model inference, post-processing, and complete synchronized prediction-call latency are recorded independently.

Platform telemetry is collected concurrently with \texttt{tegrastats} at 100-ms intervals. Recorded measurements include whole-module power, GPU utilization, unified-memory consumption, and the available thermal sensors. A 20-s idle-power measurement is collected under the same power and clock configuration before the deployment experiments, and each measured run begins after a cooldown procedure targeting approximately $50\,^{\circ}\mathrm{C}$.

Both gross and idle-subtracted energy efficiency are considered. Gross energy per processed image is obtained from mean whole-module power divided by achieved throughput. Dynamic energy additionally removes the measured idle module-power component before normalization by throughput. Unlike the reference-GPU measurements, which represent GPU-board power, the Jetson measurements represent whole-module consumption; absolute energy values are therefore interpreted within each platform rather than as direct cross-platform power comparisons.

\subsection{DeepStream Video-Analytics Configuration}
\label{sec:video_setup}

The final video experiment uses the FP16 TensorRT YOLO26s configuration selected from the embedded deployment study. APPLE MOTS test frames are retained in their original temporal order and encoded as H.264 video at 30~FPS while preserving the native $1296\times972$ frame geometry. The DeepStream processing chain consists of file input and H.264 parsing, hardware-accelerated decoding, stream multiplexing through \texttt{nvstreammux}, TensorRT detector inference through \texttt{nvinfer}, CPU-based ByteTrack association, custom motion-aware counting analytics, and the output sink.

The pipeline operates at batch size one and performs detector inference once for every input frame. The primary detector pre-cluster confidence threshold is set to 0.15 so that lower-confidence detections remain available to the second ByteTrack association stage. ByteTrack uses a high-confidence threshold of 0.25, a low-confidence threshold of 0.10, and a new-track threshold of 0.25. The track buffer is 30 frames, confidence-score fusion is enabled, and the final association matching threshold is 0.95. The latter is selected using validation sequence 0005 and subsequently frozen for test sequences 0006--0008. No ByteTrack parameter is adapted independently to an individual test sequence.

The motion-aware geometry described in Section~\ref{sec:motion_calibration} is estimated from the first 60 frames (2~s) of each video using detector--ByteTrack predictions only. These frames form the unlabeled scene-calibration interval and are excluded from the counting evaluation; counting therefore begins from frame 60. Consequently, the post-calibration counting evaluation comprises 41, 43, and 262 frames for sequences 0006, 0007, and 0008, respectively. The geometry estimation uses tracklets containing at least five observations, spanning at least four frames, and exhibiting a minimum bottom-center displacement of 12 pixels. Tracklet endpoints are estimated from three-observation endpoint windows. Robust dominant-direction estimation uses a minimum cosine compatibility of 0.50 and requires at least three supporting tracklets.

For coverage-based gate positioning, 10\% of each tracklet projection interval is removed near its endpoints, subject to a minimum margin of 5 pixels. At least three intersected calibration tracklets are required for a valid coverage estimate. Sequences 0007 and 0008 use a single full-frame motion region. Because sequence 0006 simultaneously exposes both orchard rows, a scene-level spatial partition is predefined at the horizontal image midpoint, yielding non-overlapping left and right regions. This partition is determined solely from the visible acquisition geometry and does not use ground-truth identities, crossing annotations, or test-set performance. Within each predefined region, the dominant image-plane motion and counting-gate location are estimated automatically from the unlabeled calibration tracklets.

During online counting, the bottom-center point of every ByteTrack box is used as the trajectory position. Predicted trajectory observations separated by more than seven frames are not connected for crossing detection. Crossing direction is estimated from a history of up to five recent trajectory positions, and the loose directional criterion requires motion to have a positive projection onto the calibrated image-plane direction. Each tracker identity can contribute at most one counting event across all gates.

Ground-truth crossing generation uses the identical frozen line geometry and motion-direction definition but a stricter maximum trajectory gap of three frames. This asymmetric treatment is intentional: the three-frame ground-truth limit conservatively avoids creating reference crossings by interpolating across longer annotation gaps, whereas the seven-frame prediction limit permits short interruptions in the estimated trajectories without connecting widely separated observations.

Two video-runtime modes are evaluated. In \emph{maximum-throughput} mode, sink synchronization is disabled and the pipeline processes frames as rapidly as computationally possible. In \emph{timestamp-synchronized} mode, sink synchronization preserves the configured 30-FPS presentation cadence. Quality-of-service frame dropping is disabled in both modes. The first 30 frames are excluded from runtime statistics to remove pipeline warm-up effects, and whole-module telemetry is sampled using \texttt{tegrastats} at 100-ms intervals. Power and energy measurements therefore characterize the finite duration of each evaluated video sequence rather than long-duration steady-state operation. Optional on-screen display is evaluated separately and is excluded from the principal maximum-throughput measurements.

\subsection{Evaluation Protocol}
\label{sec:evaluation_metrics}

Image-level detection is evaluated using precision, recall, mAP@50, and mAP@50:95 following the conventional COCO-style object-detection protocol \cite{lin2014microsoft}. Here, mAP@50 denotes average precision at an intersection-over-union (IoU) threshold of 0.50, whereas mAP@50:95 averages AP over IoU thresholds from 0.50 to 0.95 in increments of 0.05. For the controlled detector benchmark, results are reported as mean and sample standard deviation over the five independent training runs. Accuracy of every TensorRT engine is re-evaluated on the unchanged test partition using the corresponding fixed validation-selected checkpoint, allowing precision-induced accuracy changes to be separated from training variability.

Runtime measurements distinguish model execution from application-level processing. Backend-reported inference latency represents detector execution itself, whereas synchronized prediction latency additionally includes preprocessing, post-processing, and conversion to final detections for an image already available in memory. For the reference PyTorch image benchmark, E2E latency further includes source-image reading and decoding. Mean, median, sample standard deviation, P95, P99, and achieved throughput are retained where applicable. Resource measurements include GPU utilization, memory consumption, mean GPU-board or whole-module power, energy per processed image or frame, and maximum observed temperature.

Multi-object tracking is evaluated against the persistent APPLE MOTS identities. MOTA is reported according to the CLEAR MOT formulation \cite{bernardin2008evaluating}, while identity consistency is characterized using IDF1 \cite{ristani2016performance}. For these metrics, predicted and ground-truth boxes are associated using an IoU threshold of 0.50. Identity switches and trajectory fragmentations are additionally retained as diagnostic measures because both can directly influence downstream fruit-counting reliability.

Higher Order Tracking Accuracy (HOTA) is also reported to separate detection and association quality more explicitly \cite{luiten2021hota}. HOTA is computed using the standard set of 19 localization thresholds $\alpha=0.05,0.10,\ldots,0.95$. Sequence-level HOTA, detection accuracy (DetA), and association accuracy (AssA) are reported as averages over these thresholds following the standard HOTA evaluation convention. Reporting MOTA, IDF1, and HOTA jointly is particularly useful for the present application because a tracker can retain strong frame-level detection performance while still producing unstable fruit identities.

Counting is evaluated independently from the tracking metrics. Reference crossing events are generated from the persistent ground-truth bottom-center trajectories using the same frozen motion-aware gates and direction criteria as the deployed analytics. Thus, the reference crossing count is geometry-conditioned and represents the number of annotated fruit trajectories satisfying the predefined crossing criterion, rather than a census of all annotated fruit identities in the video. Reference trajectories separated by more than three frames are not interpolated. Predicted events use the seven-frame tracker-gap limit defined above.

Predicted and reference crossings are matched one-to-one. Events can be matched only when they belong to the same counting gate. Candidate pairs must occur within 0.20~s and within a Euclidean image-plane distance equal to 8\% of the image diagonal. These tolerances allow moderate temporal offsets in the estimated crossing instant and spatial variation in trajectory localization while restricting matches to crossings that remain close in both time and image space. Expressing the spatial tolerance relative to the image diagonal makes the criterion proportional to the frame geometry. At the configured 30~FPS and $1296\times972$ pixels, the temporal and spatial tolerances correspond to six frames and 129.6 pixels, respectively. For candidate pairs satisfying both conditions, assignment minimizes the sum of normalized temporal and spatial differences through linear-sum assignment.

Ground-truth trajectories containing a gap larger than the permitted three-frame interpolation interval can yield a geometrically plausible but temporally unsupported crossing. Such cases are retained separately as ambiguous GT-gap intervals rather than being declared reference events. A predicted crossing occurring within the same temporal and spatial tolerance of one of these ambiguous intervals is excluded from event scoring, contributing neither a true positive nor a false positive. This conservative ignore rule prevents the evaluation from penalizing a prediction when the available ground-truth trajectory does not provide sufficient temporal evidence to determine whether a reference crossing occurred.

After one-to-one matching, counting performance is summarized using event-level precision, recall, and F1 score, together with the reference and predicted crossing counts, signed and absolute count error, and relative count error. For the two-region sequence 0006, the same measures are additionally computed independently for \texttt{CountLeft} and \texttt{CountRight}; their TP, FP, FN, reference-count, and predicted-count totals are verified to sum exactly to the aggregate sequence-level evaluation.

Finally, video-runtime evaluation distinguishes computational capacity from source-rate compliance. Maximum-throughput runs report the highest sustained processing rate of the complete decoder--detector--tracker--analytics chain. Timestamp-synchronized runs are evaluated against the configured 30-FPS input cadence using input/output frame counts and the distribution of consecutive output-frame intervals. Decoder-to-tracker and decoder-to-analytics measurements represent per-frame residence latency through the pipelined DeepStream graph and are therefore reported separately from throughput; they are not interpreted as the reciprocal of the frame-processing rate.

\section{Results}
\label{sec:results}

\subsection{Image-Level Detection Performance}
\label{sec:detection_results}

Table~\ref{tab:detection_results} summarizes the test-set detection performance of YOLO26s on the four datasets. Results are reported as the mean and sample standard deviation across five independent training runs.

\begin{table}[H]
\centering
\caption{Test-set detection performance of YOLO26s across five independent training runs. Values are mean $\pm$ sample standard deviation.}
\label{tab:detection_results}

\normalsize
\setlength{\tabcolsep}{8pt}
\resizebox{\textwidth}{!}{%

\begin{tabular}{lcccc}
\toprule
\textbf{Dataset} &
\textbf{Precision} &
\textbf{Recall} &
\textbf{mAP@50} &
\textbf{mAP@50:95} \\
\midrule
MangoYOLO
& $0.9642\pm0.0045$
& $0.9617\pm0.0044$
& $0.9904\pm0.0004$
& $0.7197\pm0.0036$ \\

DeepBlueberry
& $0.7982\pm0.0136$
& $0.7684\pm0.0042$
& $0.8484\pm0.0033$
& $0.6813\pm0.0066$ \\

StrawDI\_Db1
& $0.9734\pm0.0022$
& $0.9357\pm0.0039$
& $0.9756\pm0.0017$
& $0.8656\pm0.0017$ \\

APPLE MOTS
& $0.7692\pm0.0174$
& $0.7392\pm0.0176$
& $0.8189\pm0.0232$
& $0.4957\pm0.0184$ \\
\bottomrule
\end{tabular}
}
\end{table}

Despite using the same detector architecture and training protocol, detection performance varies substantially across datasets. StrawDI\_Db1 achieves the highest mAP@50:95, reaching $0.8656\pm0.0017$, together with high precision and recall. MangoYOLO achieves the highest mAP@50 at $0.9904\pm0.0004$ and similarly high precision and recall, although its mAP@50:95 decreases to $0.7197\pm0.0036$ when performance is averaged over stricter IoU thresholds. DeepBlueberry reaches $0.6813\pm0.0066$ mAP@50:95, whereas APPLE MOTS is the most challenging task under the adopted protocol, with $0.4957\pm0.0184$.

The repeated runs are generally stable, particularly for MangoYOLO and StrawDI\_Db1. APPLE MOTS exhibits greater run-to-run variability, including standard deviations of 0.0232 for mAP@50 and 0.0184 for mAP@50:95. These results establish the multi-crop detection baseline for the subsequent deployment experiments. From this point onward, one validation-selected checkpoint per dataset is kept fixed across runtime configurations so that changes in computational performance and numerical precision are evaluated independently from training variability.

\subsection{Reference-GPU Runtime Characterization}
\label{sec:reference_gpu_results}

Table~\ref{tab:reference_gpu_results} summarizes the reference-GPU measurements, with paired values denoting FP32/FP16 execution. Three timing scopes are distinguished. \textit{Inference latency} denotes the backend-reported GPU execution time of the YOLO26s model itself. \textit{Prediction latency} is the CUDA-synchronized wall-clock duration of the complete prediction call for an image already available in memory, including preprocessing, model inference, postprocessing, and conversion of the model output to final detections, but excluding file reading and image decoding. For the PyTorch image-processing benchmark, \textit{end-to-end (E2E) latency} additionally includes reading and decoding the native-resolution source image and therefore represents the complete sequential time from image file to final detections. TensorRT experiments use images preloaded into system memory and consequently report inference and prediction latency without disk-I/O or decoding time. The TensorRT accuracy values correspond to the fixed validation-selected checkpoint used for runtime evaluation rather than the five-run averages reported in Table~\ref{tab:detection_results}.

\begin{table*}[t]
\centering
\caption{Reference-GPU runtime results on the NVIDIA GeForce RTX 5060 Ti. Paired values denote FP32/FP16 execution. Inference latency denotes model execution, while prediction latency measures the CUDA-synchronized preprocessing-to-detections path for an image already loaded in memory. PyTorch end-to-end (E2E) latency additionally includes file reading and image decoding. TensorRT throughput is derived from prediction latency. Energy is derived from GPU-board power and should therefore not be interpreted as directly comparable to the whole-module energy measurements reported for the Jetson platform.}
\label{tab:reference_gpu_results}

\small
\setlength{\tabcolsep}{3pt}

\resizebox{\textwidth}{!}{%
\begin{tabular}{lcccccccc}
\toprule
&
\multicolumn{3}{c}{\textbf{PyTorch FP32 / FP16}} &
\multicolumn{5}{c}{\textbf{TensorRT FP32 / FP16}} \\
\cmidrule(lr){2-4}
\cmidrule(lr){5-9}

\textbf{Dataset} &
\shortstack{\textbf{Inference}\\\textbf{(ms)}} &
\shortstack{\textbf{Prediction}\\\textbf{(ms)}} &
\shortstack{\textbf{E2E}\\\textbf{(ms)}} &
\shortstack{\textbf{Inference}\\\textbf{(ms)}} &
\shortstack{\textbf{Prediction}\\\textbf{(ms)}} &
\shortstack{\textbf{Throughput}\\\textbf{(img/s)}} &
\shortstack{\textbf{mAP}\\\textbf{@50:95}} &
\shortstack{\textbf{Energy}\\\textbf{(J/image)}} \\
\midrule

MangoYOLO
& 5.908 / 6.112
& 7.222 / 7.425
& 13.117 / 13.068
& 2.235 / 1.048
& 3.527 / 2.340
& 283.57 / 427.43
& 0.7203 / 0.7207
& 0.2697 / 0.1603 \\

DeepBlueberry
& 6.074 / 6.289
& 7.647 / 7.884
& 23.449 / 23.250
& 2.218 / 1.010
& 3.713 / 2.494
& 269.31 / 401.04
& 0.6705 / 0.6707
& 0.2537 / 0.0923 \\

StrawDI\_Db1
& 5.939 / 6.145
& 7.397 / 7.601
& 17.964 / 18.160
& 2.237 / 1.051
& 3.622 / 2.438
& 276.07 / 410.25
& 0.8591 / 0.8587
& 0.2719 / 0.1601 \\

APPLE MOTS
& 5.941 / 6.135
& 7.464 / 7.763
& 25.466 / 25.200
& 2.214 / 1.049
& 3.717 / 2.550
& 269.01 / 392.09
& 0.4935 / 0.4924
& 0.3000 / 0.1363 \\
\bottomrule
\end{tabular}%
}
\end{table*}

Native PyTorch inference latency is largely independent of the dataset because all images are transformed to the same $640\times640$ network input. FP32 inference ranges from 5.908 to 6.074~ms, with synchronized prediction latency between 7.222 and 7.647~ms. FP16 does not improve either quantity: inference increases slightly to 6.112--6.289~ms and prediction latency to 7.425--7.884~ms. Reduced numerical precision alone therefore provides no runtime advantage through the evaluated native PyTorch execution path.

In contrast, disk-inclusive E2E latency is strongly dataset dependent. Under FP32 it ranges from 13.117~ms for MangoYOLO to 25.466~ms for APPLE MOTS, despite the similar inference times. DeepBlueberry and APPLE MOTS incur particularly large image-loading and decoding overheads because of their higher native image resolutions. This result illustrates the distinction between model execution and complete image-processing latency: a detector with an almost constant forward-pass time can exhibit substantially different application-level performance when native image handling is included.

TensorRT substantially reduces model-execution time. FP32 inference latency decreases to 2.214--2.237~ms, while FP16 further reduces it to 1.010--1.051~ms. Relative to TensorRT FP32, this corresponds to an inference-level FP16 speedup of approximately $2.11\times$--$2.20\times$. At the prediction-call level, latency decreases from 3.527--3.717~ms to 2.340--2.550~ms, corresponding to a smaller but still substantial speedup of $1.46\times$--$1.51\times$. The difference between the two speedups reflects the contribution of preprocessing, postprocessing, and framework overhead outside the optimized TensorRT forward pass.

TensorRT throughput consequently increases from 269.01--283.57 images/s in FP32 to 392.09--427.43 images/s in FP16. The corresponding change in detection accuracy is negligible: the FP16 mAP@50:95 difference relative to TensorRT FP32 ranges from $-0.0011$ to $+0.0004$. Energy per processed image also decreases for all four datasets. Overall, these results show that the principal reference-GPU efficiency gain originates from TensorRT optimization rather than from reduced-precision PyTorch execution alone.

\subsection{Embedded Deployment on the Jetson Orin Nano Super}
\label{sec:embedded_results}

Table~\ref{tab:jetson_results} reports the deployment results on the NVIDIA Jetson Orin Nano Super. All embedded configurations operate on preloaded images; inference latency therefore represents model execution, while prediction latency represents the synchronized prediction call including preprocessing and postprocessing but excluding disk I/O. Accuracy changes are computed relative to PyTorch FP32 for the same validation-selected checkpoint.

\begin{table*}[t]
\centering
\caption{YOLO26s deployment results on the NVIDIA Jetson Orin Nano Super. The mAP change is reported relative to PyTorch FP32 for the corresponding validation-selected checkpoint. Inference latency denotes model execution, while prediction latency measures the CUDA-synchronized preprocessing-to-detections path for an image already loaded in memory, including inference and postprocessing but excluding file reading and image decoding. Throughput is derived from prediction latency. Power and energy correspond to whole-module Jetson measurements.}
\label{tab:jetson_results}

\small
\setlength{\tabcolsep}{3pt}

\resizebox{\textwidth}{!}{%
\begin{tabular}{llccccccc}
\toprule
\textbf{Dataset} &
\textbf{Configuration} &
\shortstack{\textbf{mAP}\\\textbf{@50:95}} &
\shortstack{\textbf{mAP}\\\textbf{change}} &
\shortstack{\textbf{Inference}\\\textbf{(ms)}} &
\shortstack{\textbf{Prediction}\\\textbf{(ms)}} &
\shortstack{\textbf{Throughput}\\\textbf{(img/s)}} &
\shortstack{\textbf{Power}\\\textbf{(W)}} &
\shortstack{\textbf{Energy}\\\textbf{(J/image)}} \\
\midrule

MangoYOLO
& PyTorch FP32
& 0.7228 & $0.0000$ & 29.244 & 36.671 & 27.27 & 13.06 & 0.4789 \\
& PyTorch FP16
& 0.7233 & $+0.0005$ & 29.511 & 36.795 & 27.18 & 10.73 & 0.3948 \\
& TensorRT FP32
& 0.7205 & $-0.0023$ & 12.151 & 19.464 & 51.38 & 14.73 & 0.2867 \\
& TensorRT FP16
& 0.7208 & $-0.0020$ & 6.157 & 13.412 & 74.56 & 12.30 & 0.1650 \\
& TensorRT INT8
& 0.7015 & $-0.0213$ & 6.128 & 13.415 & 74.54 & 10.63 & 0.1427 \\
\midrule

DeepBlueberry
& PyTorch FP32
& 0.6730 & $0.0000$ & 29.113 & 37.215 & 26.87 & 12.91 & 0.4805 \\
& PyTorch FP16
& 0.6719 & $-0.0012$ & 29.708 & 37.722 & 26.51 & 10.63 & 0.4008 \\
& TensorRT FP32
& 0.6704 & $-0.0026$ & 12.141 & 20.112 & 49.72 & 14.48 & 0.2912 \\
& TensorRT FP16
& 0.6710 & $-0.0021$ & 6.181 & 14.105 & 70.90 & 12.15 & 0.1713 \\
& TensorRT INT8
& 0.6364 & $-0.0366$ & 6.254 & 14.154 & 70.65 & 10.53 & 0.1490 \\
\midrule

StrawDI\_Db1
& PyTorch FP32
& 0.8643 & $0.0000$ & 28.890 & 37.051 & 26.99 & 13.03 & 0.4827 \\
& PyTorch FP16
& 0.8635 & $-0.0008$ & 29.761 & 37.879 & 26.40 & 10.63 & 0.4028 \\
& TensorRT FP32
& 0.8590 & $-0.0052$ & 12.083 & 20.159 & 49.61 & 14.40 & 0.2904 \\
& TensorRT FP16
& 0.8588 & $-0.0054$ & 6.175 & 14.212 & 70.36 & 12.01 & 0.1707 \\
& TensorRT INT8
& 0.8250 & $-0.0393$ & 6.073 & 14.102 & 70.91 & 10.42 & 0.1470 \\
\midrule

APPLE MOTS
& PyTorch FP32
& 0.4954 & $0.0000$ & 29.094 & 37.999 & 26.32 & 12.79 & 0.4860 \\
& PyTorch FP16
& 0.4947 & $-0.0007$ & 29.626 & 38.447 & 26.01 & 10.57 & 0.4066 \\
& TensorRT FP32
& 0.4935 & $-0.0019$ & 12.163 & 20.985 & 47.65 & 14.15 & 0.2969 \\
& TensorRT FP16
& 0.4923 & $-0.0031$ & 6.172 & 14.979 & 66.76 & 11.71 & 0.1754 \\
& TensorRT INT8
& 0.4716 & $-0.0238$ & 6.094 & 14.881 & 67.20 & 10.32 & 0.1536 \\
\bottomrule
\end{tabular}%
}
\end{table*}

Native PyTorch execution on the Jetson again shows little benefit from FP16. FP32 inference latency remains within 28.890--29.244~ms across the four datasets, while FP16 ranges from 29.511 to 29.761~ms. Prediction throughput therefore remains approximately 26--27 images/s for both precisions. FP16 nevertheless reduces mean module power from 12.79--13.06~W to 10.57--10.73~W, lowering energy consumption despite the absence of a latency improvement.

TensorRT produces a pronounced reduction in model-execution time. FP32 TensorRT lowers inference latency to 12.083--12.163~ms, corresponding to an approximately $2.39\times$--$2.41\times$ inference speedup relative to PyTorch FP32. Prediction latency decreases to 19.464--20.985~ms, raising throughput to 47.65--51.38 images/s and providing an application-level speedup of approximately $1.81\times$--$1.88\times$. The associated mAP@50:95 reduction remains small, ranging from 0.0019 to 0.0052.

FP16 TensorRT further reduces inference latency to 6.157--6.181~ms. Relative to PyTorch FP32, this represents an inference-level acceleration of approximately $4.68\times$--$4.75\times$. Prediction latency decreases to 13.412--14.979~ms, yielding 66.76--74.56 images/s and an application-level speedup of $2.54\times$--$2.73\times$. The smaller prediction-level speedup again reflects preprocessing and postprocessing costs that are not reduced to the same extent as model execution.

Importantly, this acceleration is obtained with only a small change in detector accuracy. TensorRT FP16 reduces mAP@50:95 by 0.0020 for MangoYOLO, 0.0021 for DeepBlueberry, 0.0054 for StrawDI\_Db1, and 0.0031 for APPLE MOTS relative to the corresponding PyTorch FP32 checkpoint. Gross energy consumption simultaneously decreases from 0.4789--0.4860~J/image under PyTorch FP32 to 0.1650--0.1754~J/image under TensorRT FP16, corresponding to an approximately 64--66\% reduction.

An additional idle-subtracted energy analysis, not tabulated in Table~\ref{tab:jetson_results}, shows the same trend. With a measured idle module power of approximately 6.10~W, dynamic energy decreases from approximately 0.253--0.257~J/image for PyTorch FP32 to 0.083--0.085~J/image for TensorRT FP16. This indicates that the reduction in gross energy is not explained solely by the fixed platform power component.

INT8 quantization provides almost no additional latency benefit over FP16. Its inference latency ranges from 6.073 to 6.254~ms, compared with 6.157--6.181~ms for FP16, while throughput remains within 67.20--74.54 images/s compared with 66.76--74.56 images/s for FP16. In contrast, the accuracy degradation is substantially larger: mAP@50:95 decreases by 0.0213--0.0393 relative to PyTorch FP32. Gross energy is further reduced to 0.1427--0.1536~J/image, approximately 12--14\% below FP16, but this additional energy saving is obtained without a meaningful throughput gain and with a considerably larger accuracy penalty.

Maximum observed temperatures remain between $50.7,^{\circ}\mathrm{C}$ and $54.1,^{\circ}\mathrm{C}$ across the evaluated Jetson configurations under the fixed power and cooling setup. Taken together, these results identify TensorRT FP16 as the most favorable configuration for the subsequent video experiment: it retains accuracy close to the FP32 baseline while substantially reducing inference latency, prediction latency, and energy consumption. INT8 provides a modest additional energy reduction but does not improve processing throughput.

\subsection{End-to-End Video Analytics on APPLE MOTS}
\label{sec:video_results}

The final experiment evaluates the complete transition from frame-level fruit detection to continuous tracking and event-based counting on the NVIDIA Jetson Orin Nano Super. The FP16 TensorRT YOLO26s configuration selected in the embedded deployment experiments was integrated with ByteTrack and the motion-aware counting procedure described in Section~\ref{sec:video_analytics}. The ByteTrack association parameters were selected using validation sequence 0005 and frozen before evaluation on the held-out APPLE MOTS sequences 0006--0008. In particular, the final configuration uses a matching threshold of 0.95. No tracker parameters or counting thresholds were subsequently adapted to the individual test sequences. The only sequence-specific scene configuration is the predefined two-region spatial partition for sequence 0006, introduced because both orchard rows are simultaneously visible; no tracker, motion-estimation, gate-placement, or event-matching parameter is tuned using the test results. For each test sequence, the first 60 frames (2~s) are used exclusively for motion-geometry calibration and are excluded from the subsequent counting evaluation. The calibration uses detector--ByteTrack trajectories only; no ground-truth identities or reference crossing events are involved in deriving the counting geometry.

\begin{figure*}[t]
    \centering
    \includegraphics[width=\textwidth]
    {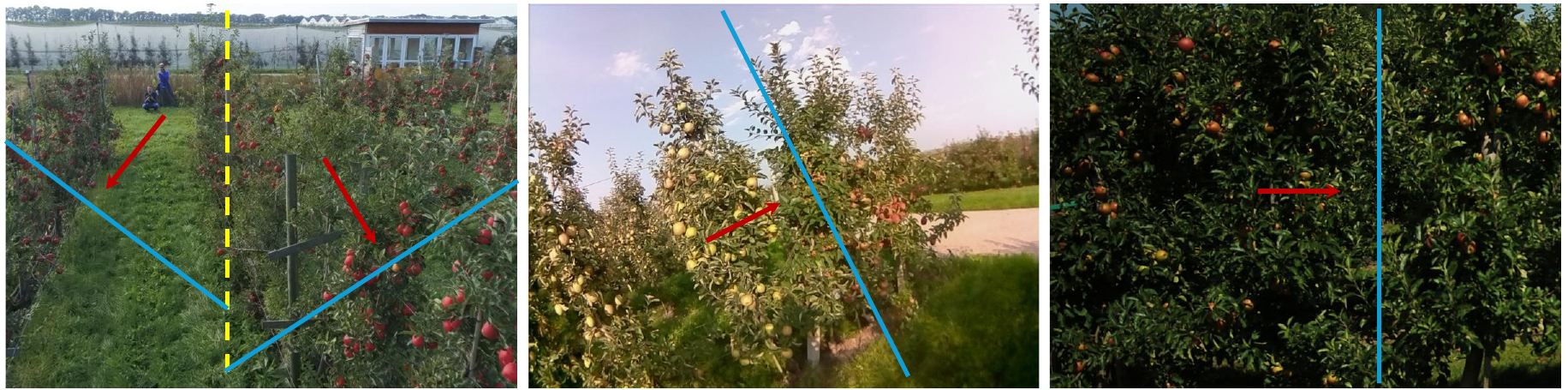}
    \caption{Representative acquisition geometries for the held-out APPLE MOTS video sequences, shown from left to right as 0006, 0007, and 0008. Sequence 0006 is acquired during forward traversal between two orchard rows and is therefore divided into two independent spatial motion regions; sequence 0007 contains an oblique reverse traversal; and sequence 0008 provides an approximately lateral view of a single orchard row. Blue arrows indicate the dominant image-plane motion estimated from ByteTrack calibration trajectories, rather than the physical camera-motion vector. Green finite lines indicate the corresponding motion-aware counting gates. The dashed partition in sequence 0006 separates the two independently modeled spatial regions.}
    \label{fig:video_geometry}
\end{figure*}

A central characteristic of this experiment is that the three test sequences represent markedly different acquisition geometries rather than repetitions of the same camera trajectory. As illustrated in Fig.~\ref{fig:video_geometry}, sequence 0006 is acquired during a forward traversal between two orchard rows, providing a wide-angle view in which fruits on the left and right sides exhibit different perspective-induced motion patterns. The scene is therefore partitioned into two non-overlapping motion regions, with independent CountLeft and CountRight gates. Sequence 0007 corresponds to an oblique reverse traversal near the end of the orchard, whereas sequence 0008 provides an approximately lateral view of a single row and exhibits substantially more uniform image-plane motion.

The arrows in Fig.~\ref{fig:video_geometry} represent the dominant \emph{image-plane motion} estimated from ByteTrack calibration trajectories and should not be interpreted as direct measurements of the physical camera motion vector. Since the fruit is stationary in the orchard while the camera moves, these vectors characterize the apparent displacement of tracked fruits in the image and therefore incorporate the effects of camera trajectory, viewpoint, depth, and perspective projection. The finite counting gates are placed approximately perpendicular to these estimated motion directions.

\subsubsection{Multi-object tracking performance}
\label{sec:video_tracking_results}

Table~\ref{tab:video_tracking_results} reports the multi-object tracking performance of the complete detector--ByteTrack pipeline. The results reveal a strong dependence on acquisition geometry. Sequence 0008 provides the highest identity consistency, achieving an IDF1 of 0.665, HOTA of 0.538, and association accuracy (AssA) of 0.681. Only 34 identity switches are observed, despite the dense appearance of fruit throughout the sequence. The HOTA decomposition further gives a detection accuracy (DetA) of 0.427, indicating that the principal advantage of this sequence is not only frame-level object availability but also the ability of ByteTrack to maintain consistent associations over time.

This behavior is consistent with the motion geometry estimated during the 60-frame calibration interval. Sequence 0008 exhibits a dominant motion angle of only $-1.26^{\circ}$ and a direction concentration of 0.9997. The tracked fruits therefore follow an almost common horizontal image-plane direction. This highly coherent motion is consistent with comparatively stable frame-to-frame spatial overlap and with the strong association performance observed for this sequence.

Sequence 0007 exhibits intermediate tracking difficulty. It achieves the highest MOTA among the three test sequences, at 0.442, while IDF1 and HOTA reach 0.557 and 0.378, respectively. Its HOTA decomposition yields $\mathrm{DetA}=0.362$ and $\mathrm{AssA}=0.397$. A total of 144 identity switches and 133 trajectory fragmentations are recorded. Although the estimated motion remains strongly coherent, with a direction concentration of 0.954, the dominant trajectory is oblique ($-18.91^{\circ}$), and the corresponding association scores are substantially lower than those obtained for the near-lateral sequence 0008.

The most challenging tracking conditions occur in sequence 0006. The final configuration obtains MOTA = 0.081, IDF1 = 0.477, and HOTA = 0.345, with $\mathrm{DetA}=0.292$ and $\mathrm{AssA}=0.411$. The sequence produces 109 identity switches and 146 trajectory fragmentations. Unlike 0007 and 0008, a single global image-motion model is not appropriate for this sequence. Within the two-region formulation, the estimated dominant motion angles are $117.54^{\circ}$ for the left region and $49.44^{\circ}$ for the right region, demonstrating that fruits on the two sides of the same frame follow substantially different apparent directions under the forward wide-angle traversal.

The difference between MOTA and the identity-oriented measures is noteworthy. Although sequence 0007 obtains a higher MOTA than 0008, sequence 0008 substantially outperforms it in IDF1, HOTA, and AssA. MOTA combines false positives, false negatives, and identity switches into a single error measure and therefore does not isolate association quality. In contrast, the strong AssA obtained for 0008 directly reflects its more stable temporal identities. For this reason, MOTA is interpreted jointly with IDF1 and the HOTA decomposition rather than as the sole indicator of tracking performance.

\begin{table*}[t]
\centering
\caption{Multi-object tracking performance on the held-out APPLE MOTS sequences using the validation-selected ByteTrack configuration. HOTA, DetA, and AssA denote the standard Higher Order Tracking Accuracy decomposition. IDSW denotes identity switches and Frag. denotes trajectory fragmentations.}
\label{tab:video_tracking_results}

\small
\setlength{\tabcolsep}{7pt}
\renewcommand{\arraystretch}{1.12}

\begin{tabular}{lccccccc}
\toprule
\textbf{Sequence} &
\textbf{MOTA} &
\textbf{IDF1} &
\textbf{HOTA} &
\textbf{DetA} &
\textbf{AssA} &
\textbf{IDSW} &
\textbf{Frag.} \\
\midrule

0006 &
0.081 &
0.477 &
0.345 &
0.292 &
0.411 &
109 &
146 \\

0007 &
\textbf{0.442} &
0.557 &
0.378 &
0.362 &
0.397 &
144 &
133 \\

0008 &
0.363 &
\textbf{0.665} &
\textbf{0.538} &
\textbf{0.427} &
\textbf{0.681} &
\textbf{34} &
\textbf{109} \\

\bottomrule
\end{tabular}
\end{table*}

\subsubsection{Event-based fruit counting}
\label{sec:video_counting_results}

The effect of temporal association quality becomes more apparent when the resulting trajectories are used for fruit counting. Reference crossing events are generated from the persistent APPLE MOTS identities using exactly the same frozen gate geometry and direction constraints as the deployed analytics. The spatial event definition is therefore identical for predictions and ground truth, while trajectory continuity is handled using the separate three-frame ground-truth and seven-frame prediction gap limits defined in Section~\ref{sec:evaluation_metrics}. Table~\ref{tab:video_counting_results} reports the resulting event-level performance.

Across the three held-out sequences, the frozen counting geometry produces 213 reference crossing events. The deployed pipeline produces 149 predicted events, of which 134 are matched to a reference event. This corresponds to 15 false-positive and 79 missed events and yields a micro-averaged event precision of 0.899, recall of 0.629, and F1 score of 0.740. The total predicted crossing count is therefore 149 compared with 213 reference crossing events, corresponding to an aggregate under-count of 64 crossing events, or 30.0\%.

Sequence 0008 again provides the strongest operational result. Of 81 reference events, 76 crossings are predicted and 63 are correctly matched. Event precision is 0.829, recall is 0.778, and F1 reaches 0.803. The sequence-level count differs from the reference by only five fruits, corresponding to a relative count error of 6.2\%.

The low count error of 0008 should nevertheless not be interpreted as equivalent to perfect event recovery. Thirteen predicted crossings are false events and 18 reference crossings are missed. The small difference between the total predicted and reference counts therefore partly results from compensating false-positive and false-negative events. This result demonstrates that a low sequence-level count error can arise from compensating false-positive and false-negative events and therefore should not be interpreted as evidence of accurate event recovery. For this reason, count error is reported jointly with event-level precision, recall, and F1.

Sequence 0007 retains perfect event precision. All 42 predicted crossings are successfully matched to reference events, resulting in no false-positive crossings. However, 26 of the 68 reference events remain undetected, limiting recall to 0.618 and producing an F1 score of 0.764. The predicted count of 42 consequently underestimates the reference count by 38.2\%. The error pattern is therefore dominated by missed crossing events rather than duplicate or unsupported counts.

Sequence 0006 remains the most difficult counting scenario. The two-region geometry yields 64 reference crossings, while the deployed pipeline produces 31 events. Of these, 29 are correct and only two are false positives, corresponding to a precision of 0.935. Recall, however, is only 0.453 because 35 reference crossings are missed. Event F1 is consequently 0.611, and the predicted total remains 51.6\% below the reference count. The combination of high precision and substantially lower recall indicates that unsupported crossing events are relatively uncommon, whereas missed events dominate the counting error. This pattern is consistent with the lower association quality and greater trajectory fragmentation observed for sequence 0006.

\begin{table*}[t]
\centering
\caption{Event-level fruit-counting performance on the held-out APPLE MOTS sequences. GT and Pred. denote reference and predicted crossing counts. TP, FP, and FN are obtained by one-to-one spatiotemporal event matching. Count error is the absolute difference between the predicted and reference crossing counts, normalized by the reference crossing count.}
\label{tab:video_counting_results}

\small
\setlength{\tabcolsep}{5pt}
\renewcommand{\arraystretch}{1.12}
\resizebox{\textwidth}{!}{%

\begin{tabular}{lcccccccccc}
\toprule
\textbf{Seq.} &
\textbf{Gates} &
\textbf{GT} &
\textbf{Pred.} &
\textbf{TP} &
\textbf{FP} &
\textbf{FN} &
\textbf{Precision} &
\textbf{Recall} &
\textbf{F1} &
\textbf{Count error} \\
\midrule

0006 &
2 &
64 &
31 &
29 &
2 &
35 &
0.935 &
0.453 &
0.611 &
51.6\% \\

0007 &
1 &
68 &
42 &
42 &
0 &
26 &
\textbf{1.000} &
0.618 &
0.764 &
38.2\% \\

0008 &
1 &
81 &
76 &
63 &
13 &
18 &
0.829 &
\textbf{0.778} &
\textbf{0.803} &
\textbf{6.2\%} \\

\midrule
\textbf{Overall} &
-- &
213 &
149 &
134 &
15 &
79 &
0.899 &
0.629 &
0.740 &
30.0\% \\

\bottomrule
\end{tabular}
}
\end{table*}

The two-region formulation additionally permits the aggregate behavior of sequence 0006 to be decomposed by orchard side. The CountLeft region contains 10 reference crossing events and produces exactly 10 predicted events. Nine are correctly matched, giving precision = 0.900, recall = 0.900, F1 = 0.900, and zero sequence-level count error for that region. In contrast, the CountRight region contains 54 reference events but produces only 21 predicted crossings. Twenty of these are correct and only one is false, yielding a high precision of 0.952 but a recall of only 0.370 and an F1 score of 0.533. The right-region count therefore underestimates the reference by 33 events, corresponding to a relative error of 61.1\%. The per-region results therefore show that the aggregate difficulty of sequence 0006 is strongly asymmetric: the same detector, tracker, and event-generation logic achieve high event recovery in the left region but remain strongly recall limited in the right region.

\subsubsection{Real-time embedded video performance}
\label{sec:video_runtime_results}

The temporal processing required for tracking and counting does not prevent real-time execution on the Jetson Orin Nano Super. As summarized in Table~\ref{tab:video_runtime_results}, maximum-throughput performance ranges from 44.96 FPS for sequence 0006 to 54.11 FPS for sequence 0008. Even the most computationally demanding video therefore provides approximately $1.50\times$ the processing capacity required for a 30-FPS input stream, while 0008 reaches approximately $1.80\times$ the configured input rate.

The computational cost of ByteTrack varies with sequence complexity. Mean tracker processing time is 18.99 ms/frame for 0006 and 19.26 ms/frame for 0007, compared with 14.59 ms/frame for 0008. The custom motion-aware counting stage contributes comparatively little additional overhead, requiring only 1.67, 2.14, and 1.41 ms/frame for sequences 0006, 0007, and 0008, respectively. In particular, introducing two independent counting regions for sequence 0006 does not compromise the real-time computational objective.

Mean module power ranges from 8.41 to 10.14 W across the three videos. The observed energy consumption is similar across these finite sequence runs, ranging from approximately 0.187 to 0.188 J per processed frame. The higher mean power observed for 0008 is offset by its higher processing rate, illustrating why instantaneous power and energy per frame should be interpreted jointly when comparing deployment configurations.

The decoder-to-tracker pipeline latency ranges from 121.0 to 156.1 ms, while decoder-to-analytics latency ranges from 122.4 to 158.3 ms. These measurements represent the residence time of an individual frame through the pipelined GStreamer/DeepStream execution graph and are therefore not the inverse of steady-state throughput. Decoding, TensorRT inference, tracking, and analytics operate concurrently on different frames, allowing the system to process more than 30 frames per second even though the latency experienced by a single frame is greater than 33.3 ms. Timing coverage is 100\% in all three sequences, with no decoder FIFO underflows, PTS fallbacks, or missing timing samples, confirming that the reported stage and pipeline latency measurements are based on complete frame-level timing records.

\begin{table*}[t]
\centering
\caption{Maximum-throughput end-to-end video-processing performance on the NVIDIA Jetson Orin Nano Super. PGIE denotes the TensorRT detector stage. Decoder-to-tracker and decoder-to-analytics values represent frame residence latency through the pipelined processing graph and should not be interpreted as inverse throughput. PGIE timing represents the detector stage within the running DeepStream pipeline and is therefore not directly equivalent to the standalone TensorRT inference latency reported in Table~\ref{tab:jetson_results}.}
\label{tab:video_runtime_results}

\small
\setlength{\tabcolsep}{5pt}
\renewcommand{\arraystretch}{1.12}
\resizebox{\textwidth}{!}{%

\begin{tabular}{lcccccccc}
\toprule
\textbf{Seq.} &
\textbf{Throughput} &
\textbf{PGIE} &
\textbf{ByteTrack} &
\textbf{Analytics} &
\textbf{Dec.$\rightarrow$Track} &
\textbf{Dec.$\rightarrow$Analytics} &
\textbf{Power} &
\textbf{Energy/frame} \\
&
\textbf{(FPS)} &
\textbf{(ms)} &
\textbf{(ms)} &
\textbf{(ms)} &
\textbf{(ms)} &
\textbf{(ms)} &
\textbf{(W)} &
\textbf{(J)} \\
\midrule

0006 &
44.96 &
18.83 &
18.99 &
1.67 &
151.60 &
153.27 &
8.41 &
0.187 \\

0007 &
47.88 &
17.52 &
19.26 &
2.14 &
156.14 &
158.28 &
9.01 &
0.188 \\

0008 &
\textbf{54.11} &
\textbf{15.97} &
\textbf{14.59} &
\textbf{1.41} &
\textbf{121.03} &
\textbf{122.44} &
10.14 &
0.187 \\

\bottomrule
\end{tabular}
}
\end{table*}

In the timestamp-synchronized experiments, the complete pipeline processed all 101/101 frames of sequence 0006, 103/103 frames of sequence 0007, and 322/322 frames of sequence 0008, with no frame loss observed at the pipeline output. The corresponding effective processing rates were 30.09, 30.08, and 30.02 FPS, respectively, closely matching the configured 30-FPS input rate. The mean inter-frame output intervals were 33.34 ms for 0006, 33.39 ms for 0007, and 33.34 ms for 0008. Temporal variation remained limited, with P95 output intervals of 36.95, 36.64, and 37.13 ms and P99 values of 39.26, 37.35, and 38.55 ms, respectively. These results confirm that the complete decoder--detector--tracker--analytics pipeline can sustain continuous 30-FPS source-rate operation on the Jetson Orin Nano Super, although a small amount of frame-to-frame scheduling jitter remains visible in the upper latency percentiles.

\section{Discussion}
\label{sec:discussion}

\subsection{Detection Performance Across Heterogeneous Fruit Datasets}
\label{sec:discussion_detection}

The multi-dataset experiment was designed to evaluate the behavior of a common lightweight detector under heterogeneous agricultural imaging conditions rather than to establish a new crop-specific state of the art. This distinction is important when interpreting the results. YOLO26s was used without architecture modifications across all four datasets, allowing differences in detection performance to be associated primarily with the characteristics of the individual detection tasks.

The resulting mAP@50:95 values span a relatively wide range, from $0.4957\pm0.0184$ for APPLE MOTS to $0.8656\pm0.0017$ for StrawDI\_Db1. This variation confirms that the same detector architecture can exhibit substantially different localization behavior depending on fruit scale, density, occlusion, image resolution, and acquisition conditions. MangoYOLO and StrawDI\_Db1 exhibit particularly strong mAP@50 values of $0.9904\pm0.0004$ and $0.9756\pm0.0017$, respectively, whereas the larger gap between mAP@50 and mAP@50:95 for several datasets indicates that accurate localization at stricter IoU thresholds remains more challenging than simply identifying the presence of fruit.

MangoYOLO provides the clearest basis for comparison with previous work because its predefined training and test partitions have been reused by multiple studies. The original MangoYOLO model reported an average precision of approximately 0.983 on the published test set \cite{koirala2019_v2deep}. More recent evaluations of mainstream YOLO architectures on the same dataset have reported mAP@50 values close to 0.99 and mAP@50:95 values of approximately 0.77 for medium-scale YOLOv8 models \cite{neupane2024developing}. The YOLO26s result obtained here ($\mathrm{mAP@50}=0.9904$ and $\mathrm{mAP@50:95}=0.7197$) is therefore competitive at the conventional 0.50 IoU threshold, while the stricter localization metric remains below that reported for some larger detector configurations. This is consistent with the objective of the present study, which prioritizes a fixed lightweight detector suitable for subsequent embedded and video deployment rather than maximizing image-level accuracy through model scaling or dataset-specific architecture modification.

Direct numerical comparison is less appropriate for DeepBlueberry and StrawDI\_Db1. The original DeepBlueberry study employed a modified Mask R-CNN formulation and reported detection mAP values of 0.759 and 0.724 at IoU thresholds of 0.50 and 0.70, respectively \cite{gonzalez2019deepblueberry}. Although the present YOLO26s experiment achieves an mAP@50 of 0.8484, differences in the data partition and evaluation procedure prevent this difference from being interpreted as a direct improvement. Similarly, StrawDI\_Db1 was introduced primarily for instance segmentation \cite{perez2020fast}; the bounding-box localization task used here is therefore not equivalent to the pixel-level segmentation benchmarks reported in the original and subsequent studies.

APPLE MOTS represents a substantially different detection environment. The sequence contains densely distributed, visually homogeneous apples and was developed specifically to study detection, segmentation, and temporal association under orchard video conditions \cite{de2022apple}. Its comparatively low mAP@50:95 of 0.4957 should therefore not be interpreted simply as a weakness of the detector. Instead, it establishes the frame-level difficulty inherited by the subsequent tracking stage. This is particularly relevant because temporal association cannot recover fruits that are consistently missed or poorly localized by the detector. 

The repeated-run experiment also provides evidence that model-training stochasticity is not negligible for every dataset. MangoYOLO and StrawDI\_Db1 show very small variation among runs, whereas APPLE MOTS exhibits considerably larger standard deviations. Reporting repeated training runs is therefore useful when comparing agricultural detection models, since a single favorable checkpoint can obscure variability attributable to model initialization and stochastic optimization.

\subsection{Embedded Inference Optimization and Accuracy--Efficiency Trade-offs}
\label{sec:discussion_deployment}

The embedded experiments demonstrate that inference backend has a substantially greater influence on computational efficiency than reduced precision within native PyTorch execution. PyTorch FP16 does not reduce latency on the Jetson Orin Nano Super, despite lowering module power. In contrast, TensorRT optimization substantially reduces both model-execution and application-level prediction latency.

FP16 TensorRT provides the most favorable overall deployment configuration. Across the four datasets, prediction throughput increases from approximately 26--27 images/s under PyTorch FP32 to 66.76--74.56 images/s under TensorRT FP16, corresponding to a $2.54\times$--$2.73\times$ application-level speedup. The associated mAP@50:95 reduction remains small (0.0020--0.0054), while gross energy consumption decreases by approximately 64--66\%. These results indicate that FP16 TensorRT provides a practically useful operating point in which substantial computational and energy savings are obtained without materially changing detector accuracy.

INT8 produces a different trade-off. Although gross energy per image is approximately 12--14\% lower than FP16, it provides essentially no additional throughput benefit. At the same time, the mAP@50:95 reduction increases to 0.0213--0.0393 relative to PyTorch FP32. Under the evaluated GPU-based TensorRT configuration, INT8 therefore offers limited practical advantage over FP16 when detection accuracy and throughput are considered jointly. This result also illustrates why reduced numerical precision should not automatically be assumed to improve real-time performance; preprocessing, postprocessing, memory movement, and hardware-specific execution characteristics can limit the benefit of further quantization.

Previous studies have demonstrated the feasibility of fruit detection and counting on embedded NVIDIA platforms. Mazzia \emph{et al.} \cite{mazzia2020real} reported real-time apple detection using YOLOv3-tiny on Jetson-class hardware, while Lyu \emph{et al.} achieved a 28-FPS detection-and-counting pipeline for green citrus on a Jetson Xavier NX \cite{lyu2022green}. More recently, improved lightweight YOLO architectures have been deployed on the Jetson Orin Nano for mango perception \cite{gu2024simultaneous}. Absolute frame-rate comparisons across these studies are not meaningful because detector architectures, image resolutions, hardware generations, preprocessing paths, and measured pipeline boundaries differ substantially. Nevertheless, they collectively demonstrate the progression of agricultural computer vision from offline GPU inference toward field-oriented embedded execution.

The present results extend this line of work by explicitly separating detector inference, synchronized prediction latency, maximum computational throughput, and source-synchronized video operation. This distinction becomes important in the complete video experiment, where detector inference is no longer the only substantial computational component.

\subsection{Tracking and Counting Under Moving-Camera Orchard Conditions}
\label{sec:discussion_tracking}

The video experiments show that satisfying the real-time computational requirement does not guarantee reliable temporal perception. All three test sequences can be processed faster than the configured 30-FPS input rate, yet their tracking and counting results differ substantially. HOTA ranges from 0.345 to 0.538 and event-level counting F1 from 0.611 to 0.803, despite use of the same detector, ByteTrack configuration, and event-generation procedure.

This result is consistent with previous research showing that tracking visually homogeneous fruits remains considerably more difficult than frame-level detection. The original APPLE MOTS study reported useful joint detection and tracking performance using TrackR-CNN and PointTrack but identified object similarity as a major challenge for maintaining persistent apple identities \cite{de2022apple}. Hernandez and Medeiros \cite{hernandez2024multi} subsequently reported an aggregate ByteTrack baseline with MOTA = 32.99\%, HOTA = 38.21\%, and IDF1 = 45.21\% under a different APPLE MOTS-based evaluation protocol. The values are not directly comparable with the present per-sequence results because the detector, sequence aggregation, and evaluation protocol differ. Nevertheless, both studies reinforce the observation that association of dense, visually similar fruits is a major source of error in moving-camera orchard video.

Tracking quality directly affects downstream counting. Villacrés \emph{et al.} \cite{villacres2023apple} compared several tracking-by-detection strategies for apple production estimation and reported an average counting error of approximately 20\% for their best configurations using YOLOv5 detections. Cascade-SORT further demonstrated that combining appearance and motion information can reduce identity switches and counting error compared with purely motion-based SORT \cite{he2022cascade}. These results suggest that richer association models can improve counting reliability, but they introduce additional computation and therefore a different accuracy--efficiency operating point from the lightweight ByteTrack configuration considered here.

The present aggregate count error of 30.0\% lies within the broad range reported for tracking-based orchard counting systems, but the sequence-level values are considerably more informative than the aggregate result. Relative count error varies from only 6.2\% for sequence 0008 to 51.6\% for sequence 0006. The variation demonstrates that a single average counting error can conceal large differences among acquisition conditions and motivates the simultaneous reporting of event-level precision, recall, and F1. Sequence 0008 further illustrates this distinction: despite a relative count error of only 6.2\%, 13 false-positive and 18 missed crossing events remain, showing that close agreement in the final count can result from compensating event-level errors.

\subsection{Influence of Acquisition Geometry}
\label{sec:discussion_geometry}

A central observation of the video experiment is that acquisition geometry strongly coincides with temporal association and counting performance. Sequence 0008, acquired from an approximately lateral viewpoint relative to a single orchard row, exhibits the highest IDF1, HOTA, and AssA values and the smallest counting error. Its calibration trajectories are almost unidirectional in the image plane, allowing fruit identities to be maintained more consistently as they approach and cross the counting gate.

Sequence 0007 represents an intermediate condition. Its oblique reverse trajectory retains sufficiently coherent apparent motion to achieve an event-level F1 of 0.764, but association accuracy is substantially lower than for 0008 and 26 of the 68 reference crossing events are missed. Sequence 0006 is considerably more challenging because the wide-angle forward traversal simultaneously observes both sides of the orchard aisle. Fruits on the two sides exhibit distinct perspective-induced motion directions, making a single global counting geometry inappropriate.

The two-region formulation addresses the geometric component of this problem by independently estimating dominant motion and counting gates for the two sides of the scene. However, the per-region results show that correcting the counting geometry does not remove the underlying association difficulty. CountLeft achieves an event F1 of 0.900 and zero count error, whereas CountRight reaches an F1 of only 0.533 and underestimates the reference count by 61.1\%. Because both regions use the same detector, tracker, and event logic, this asymmetry indicates that the temporal difficulty is strongly scene dependent.

Related studies have similarly demonstrated that orchard structure can be exploited to constrain video counting. A recent row-scale kiwifruit counting system dynamically adapted its detection region using support posts to exclude fruits from neighboring rows and reported substantial improvements in ByteTrack- and DeepSORT-based counting \cite{zhang2025row}. Although the geometric cue differs from the calibration-derived motion regions used here, both approaches support the broader conclusion that orchard counting benefits from explicitly modeling scene structure rather than applying unconstrained generic multi-object tracking.

From a practical perspective, camera trajectory should therefore be regarded as part of the design of the perception system. When tracking-based fruit counting is required, the present results suggest that approximately lateral or row-oriented acquisition may be preferable because it is associated with more coherent image-plane trajectories. Forward traversal between rows remains computationally feasible but is associated with stronger perspective variation and can require multiple spatial motion regions. Where acquisition can be controlled---for example, on a robotic platform or dedicated sensing vehicle---camera mounting and travel direction should therefore be optimized jointly with the perception pipeline.

\subsection{From Maximum Throughput to Sustained Real-Time Operation}
\label{sec:discussion_realtime}

The distinction between maximum throughput and sustained source-rate operation is another important outcome of the study. Maximum-throughput processing reaches 44.96--54.11 FPS, demonstrating that the embedded platform provides substantial computational headroom relative to the 30-FPS video input. However, unconstrained throughput alone does not establish that a pipelined video application can process a timestamped source continuously without dropping frames.

The timestamp-synchronized experiments address this issue directly. All 101, 103, and 322 frames of sequences 0006, 0007, and 0008, respectively, reach the pipeline output, with effective frame rates of 30.09, 30.08, and 30.02 FPS. Mean output intervals remain close to the nominal 33.33-ms source period, while P99 intervals remain below 40 ms. These measurements provide stronger evidence of practical real-time capability than detector-only FPS because they include decoding, TensorRT inference, ByteTrack association, motion-aware analytics, and pipeline synchronization.

The runtime decomposition also shows that temporal association is a major component of the complete workload. Mean ByteTrack time ranges from 14.59 to 19.26 ms/frame, comparable in magnitude to the 15.97--18.83 ms measured for the DeepStream detector stage, whereas the custom counting analytics add only 1.41--2.14 ms/frame. Consequently, optimization of the detector alone would provide diminishing system-level benefit once tracking becomes a comparable computational component.

Across the evaluated sequences, the observed energy consumption remains within a narrow range of approximately 0.187-–0.188 J/frame. Given the short duration of these benchmark videos, however, these measurements should be interpreted as sequence-specific observations rather than evidence of long-term energy stability across different scene complexities. For continuously operating mobile agricultural systems, such energy-per-frame measurements may be more informative than instantaneous power alone.

\subsection{Limitations and Generalizability}
\label{sec:limitations}

Several limitations should be considered when interpreting the results. First, the use of a single YOLO26s architecture is intentional and enables a controlled multi-dataset deployment study, but it does not establish that YOLO26s is the most accurate detector for any of the four crops. Larger YOLO variants, transformer-based detectors, or crop-specific architectures may provide higher image-level accuracy at increased computational cost.

Second, temporal evaluation is restricted to APPLE MOTS. The four-dataset image experiment demonstrates that the detector and deployment methodology can be applied across different fruits, but tracking and motion-aware counting are evaluated only for apples because persistent identities are available for this dataset. Validation on temporally annotated mango, blueberry, strawberry, or other orchard datasets would be required before assuming identical counting behavior across crops.

Third, the video benchmark contains only three held-out acquisition sequences. These sequences are useful because they represent markedly different camera trajectories, but they cannot span the full range of orchard layouts, camera speeds, illumination conditions, canopy structures, and sensor placements encountered in operational deployments. The observed relationship between acquisition geometry and tracking performance should therefore be validated through a larger controlled study in which viewpoint and camera trajectory are varied systematically.

Fourth, the motion-aware gate geometry is estimated during an initial calibration interval and remains fixed for the remainder of the sequence. This assumption is appropriate when camera motion remains approximately consistent but may become invalid if the platform changes direction abruptly, stops, rotates, or transitions between orchard rows. An online mechanism for detecting motion-regime changes and re-estimating the counting geometry would improve robustness in less structured trajectories. In scenes requiring multiple motion regions, the number and extent of these regions are currently specified from the visible scene layout rather than inferred automatically; future work should determine the spatial partition directly from calibration trajectories or scene structure. Because sequences 0006 and 0007 are short, the fixed 60-frame calibration interval also constitutes a substantial fraction of their duration; the reported counting results for these sequences should therefore be interpreted as post-calibration event-recovery performance rather than full-sequence counting performance.

Fifth, ByteTrack was selected because its relatively low computational overhead is well suited to embedded deployment. The tracker does not use an appearance embedding or explicit global-motion compensation. The results therefore characterize a lightweight accuracy--efficiency operating point rather than the maximum tracking accuracy achievable on APPLE MOTS. More complex association strategies may improve identity continuity, particularly in the difficult forward-motion sequences, but their computational and energy cost should be evaluated as part of the complete edge pipeline.

Finally, all embedded measurements are obtained on one NVIDIA Jetson Orin Nano Super under a fixed MAXN SUPER power mode, locked clocks, and active cooling. Absolute latency, power, and thermal values are therefore hardware- and configuration-dependent. The broader conclusions concerning the relative benefit of TensorRT FP16, the limited additional throughput from INT8, and the importance of including decoding and tracking in real-time evaluation should be tested on additional embedded platforms.

\section{Conclusion and Future Work}
\label{sec:conclusion}

This study presented a breadth-to-depth evaluation of real-time fruit perception on embedded edge hardware, progressing from multi-crop image-level detection to optimized inference and complete video-based tracking and counting. A common lightweight YOLO26s architecture was trained independently on four public datasets representing mangoes, blueberries, strawberries, and apples, allowing detection behavior to be evaluated under heterogeneous fruit sizes, densities, occlusion levels, image resolutions, and acquisition conditions without changing the detector family. Mean test mAP@50:95 ranged from 0.4957 for APPLE MOTS to 0.8656 for StrawDI\_Db1, highlighting the substantial dataset dependence of fruit-detection performance even under a common model and training protocol.

The deployment experiments demonstrated that inference backend and numerical precision strongly affect the practical efficiency of the detector. On the NVIDIA Jetson Orin Nano Super, TensorRT FP16 achieved 66.76--74.56 images/s at 13.41--14.98~ms synchronized prediction latency, while reducing mAP@50:95 by only 0.0020--0.0054 relative to PyTorch FP32. Gross energy consumption decreased by approximately 64--66\%. INT8 provided a further modest energy reduction but essentially no additional throughput benefit and introduced a substantially larger accuracy penalty. Under the evaluated hardware and software configuration, FP16 TensorRT therefore provided the most favorable accuracy--latency--energy trade-off.

The selected FP16 detector was subsequently integrated with hardware-accelerated decoding, ByteTrack multi-object tracking, and calibration-derived motion-aware counting in an NVIDIA DeepStream pipeline. The complete system achieved a maximum throughput of 44.96--54.11 FPS and sustained the configured 30-FPS video rate on all evaluated sequences without output-frame loss. These results show that real-time computational capacity can be achieved on compact embedded hardware even after temporal association and counting analytics are included.

The temporal evaluation, however, also demonstrated that computational real-time performance alone is insufficient to guarantee accurate orchard video analytics. Across the held-out video sequences, Higher Order Tracking Accuracy (HOTA) ranged from 0.345 to 0.538, while event-level counting F1 ranged from 0.611 to 0.803 and relative count error from 6.2\% to 51.6\%. The strongest performance was obtained for an approximately lateral single-row acquisition, whereas forward wide-angle traversal between orchard rows was associated with substantially more difficult association conditions. The multi-region motion-aware formulation corrected the counting geometry for this scene but could not eliminate the underlying tracking limitations. These findings indicate that camera trajectory, viewpoint, and orchard-row geometry should be treated as integral components of the perception-system design rather than as incidental properties of the input video.

Future work should therefore address both temporal perception and acquisition design. First, the current calibration-derived counting geometry remains fixed after the initial calibration interval. An online mechanism for detecting changes in the dominant image-plane motion and dynamically re-estimating spatial regions and counting gates would enable operation under camera turns, speed changes, stops, and transitions between orchard rows. Second, lightweight association methods incorporating selective appearance information or explicit camera/global-motion compensation should be investigated to reduce identity fragmentation while preserving the real-time and energy constraints of the embedded platform.

A further priority is extending temporal evaluation beyond apples. The multi-crop results demonstrate that the detection and deployment framework is applicable to different fruits, but quantitative tracking and counting were evaluated only where persistent video-level identities were available. Temporally annotated datasets for additional crops, together with controlled experiments that systematically vary camera orientation, speed, row distance, and viewing angle, would allow the relationship between acquisition geometry and temporal accuracy to be quantified more rigorously. Finally, evaluation on additional embedded platforms and power configurations would help determine which of the observed TensorRT and energy-efficiency trends generalize beyond the NVIDIA Jetson Orin Nano Super.

Overall, the results support a system-level view of edge-based agricultural vision: practical performance depends not only on detector accuracy or inference speed, but on the interaction among model optimization, temporal association, scene geometry, acquisition strategy, and embedded computing constraints.

\bibliographystyle{elsarticle-num} 
\bibliography{references}

\end{document}